\documentclass[a4paper,10pt]{article}

\usepackage{geometry}

\usepackage{natbib}
\usepackage{hyperref}
\hypersetup{colorlinks=true, linkcolor=blue, citecolor=blue, urlcolor=magenta}
\usepackage{amsmath,amssymb,amsthm}
\usepackage{bm}
\usepackage{booktabs}
\usepackage{subfig}
\usepackage{tikz}       

\newtheorem{thm}{Theorem}

\newtheorem{fact}[thm]{Fact}
\newtheorem{dfn}{Definition}

\newcommand{\refeq}[1]{\eqref{eq:#1}} 
\newcommand{\reffig}[1]{Figure~\ref{fig:#1}}

\newcommand{\refapp}[1]{Appendix~\ref{app:#1}}
\newcommand{\refsec}[1]{Section~\ref{sec:#1}}

\newcommand{\refthm}[1]{Theorem~\ref{thm:#1}}

\newcommand{\reffact}[1]{Fact~\ref{fact:#1}}

\newcommand{\xx}{\bm{x}}
\newcommand{\yy}{\bm{y}}
\newcommand{\zz}{\bm{z}}
\newcommand{\uu}{\bm{u}}
\newcommand{\nn}{\bm{n}}
\newcommand{\cx}{\widetilde{\bm{x}}}

\newcommand{\mm}{\bm{m}}
\newcommand{\eps}{{\bm{\varepsilon}}}
\renewcommand{\aa}{{\bm{a}}}
\newcommand{\bb}{{\bm{b}}}
\newcommand{\cc}{{\bm{c}}}
\newcommand{\ggamma}{{\bm{\gamma}}}
\newcommand{\ttheta}{{\bm{\theta}}}
\newcommand{\oomega}{{\bm{\omega}}}

\renewcommand{\gg}{{\bm{g}}}
\newcommand{\ff}{{\bm{f}}}
\newcommand{\hh}{{\bm{h}}}

\newcommand{\RR}{\mathbb{R}}
\newcommand{\BB}{\mathbb{B}}

\newcommand{\Sph}{\mathbb{S}}
\newcommand{\spW}{\mathcal{P}_2}
\newcommand{\spQ}{\mathcal{Q}}
\newcommand{\metW}{\mathfrak{g}_2}
\newcommand{\sploc}{L^1_{\mathrm{loc}}}
\newcommand{\spcomp}{C_c^\infty}

\newcommand{\EE}{\mathbb{E}}
\newcommand{\dd}{\mathrm{d}}
\newcommand{\vol}{\mathsf{vol}}

\newcommand{\proofhere}{\noindent{\bf Proof} \ }

\newcommand{\wdot}{\,\cdot\,}
\newcommand{\ind}{\mathbf{1}}
\newcommand{\tr}{\mathsf{tr}\,}
\newcommand{\id}{\mathsf{id}}
\newcommand{\diag}{\mathsf{diag}\,}
\newcommand{\grad}{\mathsf{grad}\,}

\newcommand{\almost}{\mathsf{a.e.}\,}
\newcommand{\rad}{\mathrm{Rad}}

\newcommand{\data}{\mu}
\newcommand{\noise}{\nu}
\newcommand{\cdae}{{\bm{\varphi}}}
\newcommand{\dec}{{\mathsf{dec}}}
\newcommand{\enc}{{\mathsf{enc}}}
\newcommand{\daenet}{\mathsf{dae}}
\newcommand{\ent}{H}
\newcommand{\free}{F}
\newcommand{\ground}{M}
\newcommand{\hidden}{H}

\newcommand{\prob}{\mathcal{P}}
\newcommand{\vf}{\bm{v}}
\newcommand{\potential}{\nabla V}
\newcommand{\func}{\bm{h}}

\newcommand{\tdata}{\widetilde{\data}}
\newcommand{\tdec}{\widetilde{\dec}}
\newcommand{\tenc}{\widetilde{\enc}}
\newcommand{\tdaenet}{\widetilde{\daenet}}
\newcommand{\tground}{\widetilde{\ground}}
\newcommand{\thidden}{\widetilde{\hidden}}
\newcommand{\tp}{{\widetilde{p}}}
\newcommand{\tzz}{\widetilde{\zz}}

\newcommand{\tnabla}{\widetilde{\nabla}}

\newcommand{\tj}{{\widetilde{\textit{\j}}}}
\newcommand{\taa}{\widetilde{\aa}}
\newcommand{\tbb}{\widetilde{\bb}}
\newcommand{\tb}{\widetilde{b}}

\newcommand{\tcc}{\widetilde{\cc}}
\newcommand{\tc}{\widetilde{c}}
\newcommand{\tL}{\widetilde{L}}
\newcommand{\tD}{\widetilde{D}}
\newcommand{\tV}{\widetilde{V}}
\newcommand{\tff}{\widetilde{\ff}}
\newcommand{\tu}{\widetilde{u}}

\newcommand{\tgamma}{\widetilde{\gamma}}

\newcommand{\cmap}{{\bm{\varphi}}}
\newcommand{\dmap}{{\bm{g}}}
\newcommand{\lmap}{\bm{\psi}}

\newcommand{\heat}{\mathrm{\phi}}
\newcommand{\dt}{\Delta t}
\newcommand{\tbound}{T}

\newcommand{\emb}[1]{\textbf{#1}}%
\begin{document}

\title{Transport Analysis of Infinitely Deep Neural Network}

\author{Sho Sonoda\thanks{RIKEN AIP} \and Noboru Murata\thanks{Waseda University}}


\maketitle

\begin{abstract}%
We investigated the feature map inside deep neural networks (DNNs) by tracking the transport map. We are interested in the \emph{role of depth}---why do DNNs perform better than shallow models?---and the \emph{interpretation} of DNNs---what do intermediate layers do?
Despite the rapid development in their application, DNNs remain analytically unexplained because the hidden layers are nested and the parameters are not faithful.
Inspired by the \emph{integral representation} of shallow NNs, which is the continuum limit of the width, or the hidden unit number,
we developed the \emph{flow representation} and \emph{transport analysis} of DNNs. The flow representation is the continuum limit of the depth, or the hidden layer number, and it is specified by an ordinary differential equation (ODE) with a vector field. We interpret an ordinary DNN as a \emph{transport map} or an Euler broken line approximation of the flow. Technically speaking, a dynamical system is a natural model for the nested feature maps. In addition, it opens a new way to the coordinate-free treatment of DNNs by avoiding the redundant parametrization of DNNs. Following \emph{Wasserstein geometry}, we analyze a flow in three aspects: dynamical system, continuity equation, and Wasserstein gradient flow. A key finding is that we specified a series of transport maps of the \emph{denoising autoencoder} (DAE), which is a cornerstone for the development of deep learning. Starting from the shallow DAE, this paper develops three topics: the transport map of the deep DAE, the equivalence between the stacked DAE and the composition of DAEs,
and the development of the double continuum limit or the integral representation of the flow representation.
As partial answers to the research questions, we found that deeper DAEs converge faster and the extracted features are better; in addition, a deep Gaussian DAE transports mass to decrease the Shannon entropy of the data distribution.
We expect that further investigations on these questions lead to the development of an interpretable and principled alternatives to DNNs.
\end{abstract}


\section{Introduction}
Despite the rapid development in their application, deep neural networks (DNN) remain analytically unexplained.
We are interested in
the \emph{role of depth}---\emph{why do DNNs perform better than shallow models?}---and the \emph{interpretation} of DNNs---\emph{what do intermediate layers do?}
To the best of our knowledge,
thus far, traditional theories, such as the statistical learning theory \citep{Vapnik1998}, have not succeeded in completely answering the above questions \citep{Zhang2018}.
Existing DNNs lack interpretability; hence, a DNN is often called a \emph{blackbox}.
In this study, we propose the \emph{flow representation} and \emph{transport analysis} of DNNs, which provide us with insights into why DNNs can perform better and facilitate our understanding of what DNNs do.
We expect that these lines of study lead to the development of an interpretable and principled alternatives to DNNs.

Compared to other \emph{shallow models}, such as kernel methods \citep{Shawe-Taylor2004} and ensemble methods \citep{Schapire2012}, DNNs have at least two specific technical issues: the \emph{function composition} and the \emph{redundant and complicated parametrization}.
First, a DNN is formally a composite $\gg_L \circ \cdots \circ \gg_0$ of intermediate maps $\gg_\ell \, (\ell = 0, \ldots, L)$. Here, each $\gg_\ell$ corresponds to the $\ell$-th hidden layer.
Currently, our understanding of learning machines is based on \emph{linear algebra}, i.e., the \emph{basis and coefficients} \citep{Vapnik1998}.
Linear algebra is compatible with shallow models because a shallow model is a linear combination of basis functions.
However, it has poor compatibility with deep models because the function composition $(\ff, \gg) \mapsto \ff \circ \gg$ is not assumed in the standard definition of the linear space.
Therefore, we should move to spaces where the function composition is defined, such as monoids, semigroups, and \emph{dynamical systems}.
Second, the standard parametrization of the NN, such as $\gg_\ell(\xx) = \sum_{j=1}^p \cc_j^\ell \sigma( \aa_j^\ell \cdot \xx - b_j^\ell )$,  is redundant because there exist different sets of parameters that specify the same function, which causes technical problems, such as local minima.
Furthermore, it is complicated because the interpretation of parameters is usually impossible, which results in the blackbox nature of DNNs.
Therefore, we need a new parametrization that is concise in the sense that different parameters specify different functions and simple in the sense that it is easy to understand.

For shallow NNs, the \emph{integral representation theory} \citep{Murata1996,Candes.PhD,Sonoda2015} provides a concise and simple reparametrization.
The integral representation is derived by a continuum limit of the \emph{width} or the number of hidden units.
Owing to the \emph{ridgelet transform} or a pseudo-inverse operator of the integral representation operator, it is concise and simple (see \refsec{intrep.intro} for further details on the ridgelet transform).
Furthermore, in the integral representation, we can compute the parameters of the shallow NN that attains the \emph{global minimum} of the backpropagation training \citep{Sonoda2018}.
In the integral representation, thus far, the shallow NNs is no longer a blackbox, and the training is principled.
However, the integral representation is again based on linear algebra, the scope of which does not include DNNs.

Inspired by the integral representation theory,
we introduced the \emph{flow representation} and developed the \emph{transport analysis} of DNNs.
The flow representation is derived by a continuum limit of the \emph{depth} or the number of hidden layers.
In the flow representation, we formulate a DNN as a flow of an ordinary differential equation (ODE) $\dot{\xx_t} = \vf_t( \xx_t )$ with vector field $\vf_t$.
In addition, we introduced the \emph{transport map} by which we call a discretization $\xx \mapsto \xx + \ff_t(\xx)$ of the flow.
Specifically, we regard the intermediate map $\gg : \RR^m \to \RR^n$ of an ordinary DNN as a transport map that transfers the mass at $\xx \in \RR^m$ toward $\gg(\xx) \in \RR^n$.
Since the flow and transport map are independent of coordinates, they enable us the coordinate-free treatment of DNNs.
In the transport analysis, following \emph{Wasserstein geometry} \citep{Villani2009},
we track a flow by analyzing the three profiles of the flow: \emph{dynamical system}, \emph{pushforward measure}, and \emph{Wasserstein gradient flow} \citep{Ambrosio2008} (see \refsec{quick} for further details).

\begin{figure}[t]
	\centering
	\includegraphics[width=0.5\textwidth]{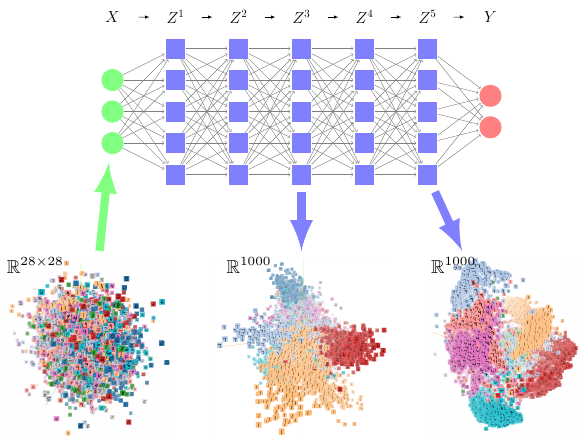}
	\caption{Mass transportation in a deep neural network that classifies images of digits.
	In the final hidden layer, the feature vectors have to be linearly separable because the output layer is just a linear classifier.
	Hence, through the network, the same digits gradually accumulate and different digits gradually separate.}
	\label{fig:tmap00}
\end{figure}

We note that when the input and the output differ in dimension, i.e., $m \neq n$,
we simply consider that both the input space and the output space are embedded in a common high-dimensional space.
As a composite of transport maps leads to another transport map, the transport map has compatibility with deep structures.
In this manner, transportation is a universal characteristic of DNNs.
For example, let us consider a digit recognition problem with DNNs.
We can expect the feature extractor in the DNN to be a transport map that separates the feature vectors of different digits,
similar to the separation of \emph{oil and water} (see \reffig{tmap00} for example).
At the time of the initial submission in 2016, the flow representation seemed to be a novel viewpoint of DNNs.
At present, it is the mainstream of development.
For example, two important DNNs---residual network (ResNet) \citep{He2015} and generative adversarial net (GAN) \citep{Goodfellow2014}---are now considered to be transport maps (see \refsec{relatedwork} for a more detailed survey).
Instead of directly investigating DNNs in terms of the redundant and complex parametrization, we perform transport analysis associated with the flow representation.
We consider that the flow representation is potentially concise and simple because the flow is independent of parametrization, and it is specified by a single vector field $\vf$.

In this study, we demonstrate transport analysis of the \emph{denoising autoencoder (DAE)}.
The DAE was introduced by \citet{Vincent2008} as a heuristic modification to enhance the robustness of the traditional autoencoder.
The traditional autoencoder is an NN that is trained as an identity map $\gg(\xx) = \xx$.
The hidden layer of the network is used as a feature map, which is often called the ``code'' because the activation pattern appears to be random, but it surely encodes some information about the input data.
On the other hand, the DAE is an NN that is trained as a ``denoising'' map
$\gg(\cx) \approx \xx$
of deliberately corrupted inputs $\cx$.
The DAE is a cornerstone for the development of deep learning or representation learning \citep{Bengio.replearn}.
Although the \emph{corrupt and denoise} principle is simple, it is successful and has inspired many representation learning algorithms (see \refsec{dae.intro} for example). Furthermore, we investigate \emph{stacking} \citep{Bengio2006} of DAEs. Because \emph{stacked DAE} \citep{Vincent2010} runs DAEs on the codes in the hidden layer, it has been less investigated, so far.

The key finding is that when the corruption process is additive, i.e., $\cx=\xx + \eps$ with some noise $\eps$,
then the DAE $\gg$ is given by the sum of the traditional autoencoder $\cx \mapsto \cx$ and a certain denoising term $\cx \mapsto \ff_t(\cx)$ parametrized by noise variance $t$:
\begin{align}
&\gg_t(\cx) = \cx + \ff_t(\cx). \label{eq:dae}
\end{align}
From the statistical viewpoint, this equation is reasonable because the DAE amounts to an estimation problem of the mean parameter.
Obviously, \refeq{dae} is a transport map because the denoising term $\ff_t$ is a displacement vector from the origin $\cx$ and the noise variance $t$ is the transport time.
Starting from the shallow DAE, this paper develops three topics: the transport map of the deep DAE, the equivalence between the stacked DAE and the composition of DAEs, 
and the development of the double continuum limit, or the integral representation of the flow representation.

\subsection{Contributions of This Study}

In this paper, we introduce the flow representation of DNNs and develop the transport analysis of DAEs.
The contributions of this paper are listed below.

\begin{itemize}
\item We introduced the flow representation, which can avoid the redundancy and complexity of the ordinary parametrization of DNNs.
\item We specified the transport maps of shallow, deep, and infinitely deep DAEs, and provided their statistical interpretations. The shallow DAE is an estimator of the mean, and the deep DAE transports data points to decrease the Shannon entropy of the data distribution. According to analytic and numerical experiments, we showed that deep DAEs can extract much more information than shallow DAEs.
\item We proved the equivalence between the stacked DAE and the composition of DAEs. Because of the peculiar construction, it is difficult to formulate and understand stacking. Nevertheless, by tracking the flow, we succeeded in formulating the stacked DAE. Consequently, we can interpret the effect of the \emph{pre-training} as a regularization of hidden layers.
\item We provided a new direction for the mathematical modeling of DNNs: the double continuum limit or the integral representation of the flow representation. We presented some examples of the double continuum limit of DAEs.
In the integral representation, the shallow NNs is no longer a blackbox, and the training is principled.
We consider that further investigations on the double continuum limit lead to the development of an interpretable and principled alternatives to DNNs.
\end{itemize}

\subsection{Related Work} \label{sec:relatedwork}

\subsubsection{Why Deep?}
Before the success of deep learning, traditional theories were skeptical of the depth concept.
According to approximation theory, (not only NNs but also) various shallow models can approximate any function \citep{Pinkus2005}.
According to estimation theory, various shallow models can attain the minimax optimal ratio \citep{Tsybakov2009}.
According to optimization theory, the depth does nothing but increase the complexity of loss surfaces unnecessarily \citep{Boyd2004}.
In reality, of course, DNNs perform overwhelmingly better than shallow models.
Thus far, the learning theory has not succeeded in explaining the gap between theory and reality \citep{Zhang2017}.

In recent years, these theories have changed drastically.
For example, many authors claim that the depth increases the expressive power in the exponential order while the width does so in the polynomial order \citep{Telgarsky2016,Eldan2016,Cohen2016,Yarotsky2017},
and that DNNs can attain the minimax optimal ratio in wider classes of functions \citep{Schmid-Hieber2017,Imaizumi2018}.
Radical reviews of the shape of loss surfaces \citep{Dauphin2014,Choromanska2015a,Kawaguchi2016,Soudry2016}, the implicit regularization by stochastic gradient descent \citep{Neyshabur.PhD}, and the acceleration effect by over-parametrization \citep{Nguyen2017,Arora2018} are ongoing.
Besides the recent trends toward the rationalization of deep learning, neutral yet interesting studies have been published \citep{Ba2014,Lin2017,Poggio2017}.
In this study, we found that deep DAEs converge faster and that the extracted features are different from each other.

\subsubsection{What Do Deep Layers Do?}
Traditionally, DNNs are said to construct the hierarchy of meanings \citep{Hinton1989}.
In convolutional NNs for image recognition, such hierarchies are empirically observed \citep{Lee2010,Krizhevsky2012,Zeiler2014}.
The hierarchy hypothesis seems to be acceptable, but it lacks explanations as to how the hierarchy is organized.

\citet{Taigman2014} reported an interesting phenomenon whereby the activation patterns in the hidden layers change by gradation from face-like patterns to codes.
\begin{figure}[t]
    \centering
    \includegraphics[width=\textwidth]{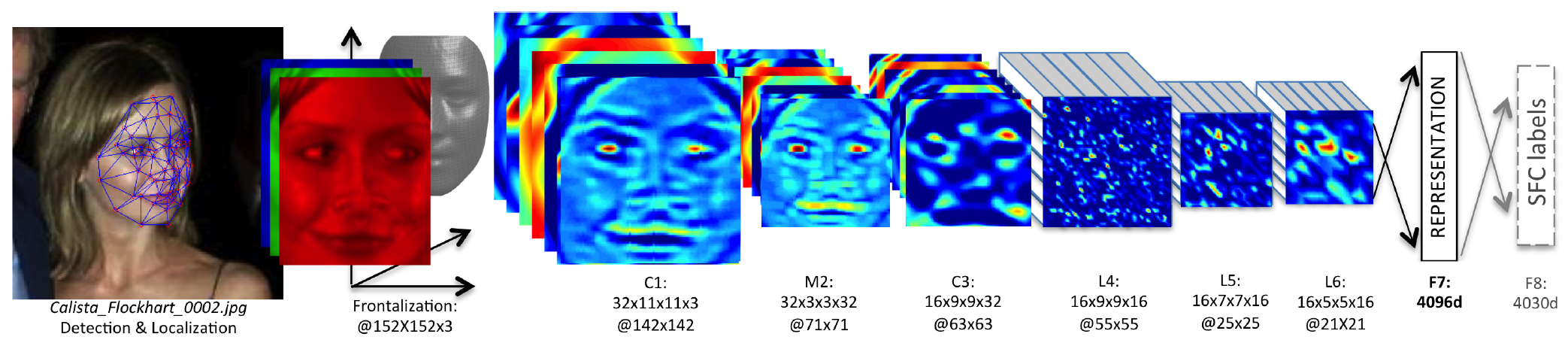}
    \caption{The activation patterns in DeepFace gradually changes \citep{Taigman2014}.}
    \label{fig:deepface}
\end{figure}
Inspired by \reffig{deepface}, we came up with the idea of regarding the activation pattern as a coordinate and the depth as the transport time.

\subsubsection{Flow Inside Neural Networks}

At the time of the initial submission in 2016, the flow representation, especially the continuum limit of the depth and collaboration with Wasserstein geometry, seemed to be a novel viewpoint of DNNs.
At present, it is the mainstream of development.

\citet{Alain2014} was the first to derive a special case of \refeq{dae}, which motivated our study.
Then, \citet{Alain2015} developed the generative model as a probabilistic reformulation of DAE.
The generative model was a new frontier at that time; now, it is widely used in variational autoencoders \citep{Kingma2014a}, generative adversarial nets (GANs) \citep{Goodfellow2014}, minimum probability flows \citep{Sohl-Dickstein2015}, and normalizing flows \citep{Rezende2015}.
Generative models have high compatibility with transport analysis because they are formulated as Markov processes.
In particular, the \emph{generator} in GANs is exactly a transport map because it is a change-of-distribution $\gg : M \to N$ from a normal distribution to a data distribution.
From this viewpoint, \citet{Arjovsky2017} succeeded in stabilizing the training process of GANs by introducing Wasserstein geometry.

The \emph{skip connection} in the residual network (ResNet) \citep{He2015} is considered to be a key structure for training a super-deep network with more than $1,000$ layers.
Formally, the skip connection is a transport map because it has an expression $\gg(\xx) = \xx + \ff(\xx)$. 
From this viewpoint, \citet{Nitanda2018} reformulated the ResNet as a functional gradient and estimated the generalization error,
and \citet{Lu2018a} unified various ResNets as ODEs.
In addition, \citet{Chizat2018} proved the global convergence of stochastic gradient descent (SGD) using Wasserstein gradient flow.
Novel deep learning methods have been proposed by controlling the flow \citep{Ioffe2015, Gomez2017, Li2018b}.

We remark that in shrinkage statistics, the expression of the transport map $\xx + \ff(\xx)$ is known as Brown's representation of the posterior mean \citep{George2006}.
\citet{Liu2016} analyzed it and proposed a Bayesian inference algorithm, apart from deep learning.

\subsection{Background}

\subsubsection{Denoising Autoencoders} \label{sec:dae.intro}

The \emph{denoising autoencoder (DAE)} is a fundamental model for representation learning,
the objective of which is to capture a good representation of the data. 
\citet{Vincent2008} introduced it as a heuristic modification of traditional autoencoders for enhancing robustness.
In the setting of traditional autoencoders, we train an NN as an identity map $\xx \mapsto \xx$ and extract the hidden layer to obtain the so-called ``code.''
On the other hand, the DAE is trained as a denoising map $\cx \mapsto \xx$ of deliberately corrupted inputs $\cx$.
Although the \emph{corrupt and denoise} principle is simple, it has inspired many next-generation models.
In this study, we analyze DAE variants such as shallow DAE, deep DAE (or composition of DAEs), infinitely deep DAE (or continuous DAE), and stacked DAE. Stacking \citep{Bengio2006} was proposed in the early stages of deep learning, and it remains a mysterious treatment because it runs DAEs on codes in the hidden layer.

The theoretical justifications and extensions follow from at least five standpoints: manifold learning \citep{Rifai2011, Alain2014}, generative modeling \citep{Vincent2010, Bengio2013, Bengio2014}, infomax principle \citep{Vincent2010}, learning dynamics \citep{Erhan2010}, and score matching \citep{Vincent2011}.
The first three standpoints were already mentioned in the original paper \citep{Vincent2008}.
According to these standpoints, a DAE extracts one of the following from the data set: a manifold on which the data are arranged (manifold learning); the latent variables, which often behave as nonlinear coordinates in the feature space, that generate the data (generative modeling); a transformation of the data distribution that maximizes the mutual information (infomax); good initial parameters that allow the training to avoid local minima (learning dynamics); or the data distribution (score matching).
A turning point appears to be the finding of the score matching aspect \citep{Vincent2011}, which reveals that score matching with a special form of the energy function coincides with a DAE. Thus, a DAE is a density estimator of the data distribution $\data$. In other words, it extracts and stores information as a function of $\data$.
Since then, many researchers have avoided stacking deterministic autoencoders and have developed generative density estimators \citep{Bengio2013, Bengio2014} instead.

\subsubsection{Integral Representation Theory and Ridgelet Analysis} \label{sec:intrep.intro}
The flow representation is inspired by the integral representation theory \citep{Murata1996, Candes.PhD, Sonoda2015}.

The integral representation
\begin{align}
S[\gamma](\xx)=\int \gamma(\aa,b) \sigma( \aa \cdot \xx - b ) \dd \lambda(\aa,b)
\end{align}
is a continuum limit of a shallow NN $g_p(\xx) = \sum_{j=1}^p c_j \sigma( \aa_j \cdot \xx - b_j )$ as the hidden unit number $p \to \infty$.
In $S[\gamma]$, every possible nonlinear parameter $(\aa, b)$ is ``integrated out,'' and only linear parameters $c_j$ remain as a coefficient function $\gamma(\aa,b)$.
Therefore, we do not need to select which $(\aa,b)$'s to use, which amounts to a non-convex optimization problem. Instead, the coefficient function $\gamma(\aa,b)$ automatically selects the $(\aa, b)$'s by weighting them. Similar reparametrization techniques have been proposed for Bayesian NNs \citep{Neal1996} and convex NNs \citep{LeRoux2006,Bach2014}.
Once a coefficient function $\gamma$ is given, we can obtain an ordinary NN $g_p$ that approximates $S[\gamma]$ by numerical integration.
We also remark that the integral representation $S[\gamma_p]$ with a singular coefficient $\gamma_p := \sum_{j=1}^p c_j \delta_{(\aa_j,b_j)}$ leads to an ordinary NN $g_p$.

The advantage of the integral representation is that the solution operator---the \emph{ridgelet transform}---to the integral equation $S[\gamma]=f$ and the optimization problem of $L[\gamma] := \| S[\gamma] - f \|^2 + \beta \| \gamma \|^2$ is known.
The ridgelet transform with an admissible function $\rho$ is given by
\begin{align}
R[f](\aa,b) := \int_{\RR^m} f(\xx) \overline{\rho(\aa \cdot \xx - b)} \dd \xx.
\end{align}
The integral equation $S[\gamma] = f$ is a traditional form of learning, and the ridgelet transform $\gamma = R[f]$ satisfies $S[\gamma] = S[R[f]] = f$ \citep{Murata1996, Candes.PhD, Sonoda2015}.
The optimization problem of $L[\gamma]$ is a modern form of learning, and a modified version of the ridgelet transform gives the global optimum \citep{Sonoda2018}.
These studies imply that a shallow NN is \emph{no longer a blackbox} but a ridgelet transform of the data set.
Traditionally, the integral representation has been developed to estimate the approximation and estimation error bounds of shallow NNs $g_p$ \citep{Barron1993,Kurkova2012,Klusowski2017,Klusowski2016,Suzuki2017}.
Recently, the numerical integration methods for $R[f]$ and $S[R[f]]$ were developed \citep{Candes.PhD,Sonoda2014,Bach2015} with various $f$, including the MNIST classifier.
Hence, by computing the ridgelet transform of the data set, we can obtain the global minimizer without gradient descent.

Thus far, the integral representation is known as an efficient reparametrization method
to facilitate understanding of the hidden layers,
to estimate the approximation and estimation error bounds of shallow NNs,
and to calculate the hidden parameters.
However, it is based on linear algebra, i.e., it starts by regarding $c_j$ and $\sigma(\aa_j \cdot \xx - b_j)$ as coefficients and basis functions,
respectively. Therefore, the integral representation for DNNs is not trivial at all.

\subsubsection{Optimal Transport Theory and Wasserstein Geometry} \label{sec:quick.intro}

The optimal transport theory \citep{Villani2009} originated from the practical requirement in the 18th century to transport materials at the minimum cost.
At the end of the 20th century, it was transformed into \emph{Wasserstein geometry,} or the geometry on the space of probability distributions.
Recently, Wasserstein geometry has attracted considerable attention in statistics and machine learning.
One of the reasons for its popularity is that the \emph{Wasserstein distance} can capture the difference between two singular measures, whereas the traditional Kullback-Leibler distance cannot \citep{Arjovsky2017}. Another reason is that it gives a unified perspective on a series of function inequalities, including the concentration inequality. Computation methods for the Wasserstein distance and Wasserstein gradient flow have also been developed \citep{Peyre2018,Nitanda2018,Zhang2018}.
In this study, we employ \emph{Wasserstein gradient flow} \citep{Ambrosio2008} for the characterization of DNNs.

Given a density $\mu$ of materials in $\RR^m$, a density $\nu$ of final destinations in $\RR^m$, and a cost function $c : \RR^m \times \RR^m \to \RR$ associated with the transportation,
under some regularity conditions, there exist some optimal transport map(s) $\gg : \RR^m \to \RR^m$ that attain the minimum transportation cost.
Let $W(\mu,\nu)$ denote the minimum cost of the transportation problem from $\mu$ to $\nu$.
Then, it behaves as the distance between two probability densities $\mu$ and $\nu$, and it is called the \emph{Wasserstein distance}, which is the start point of Wasserstein geometry. 

When the cost function $c$ is given by the $\ell^p$-distance, i.e., $c(\xx,\yy) = |\xx - \yy|_p$, the corresponding Wasserstein distance is called the $L^p$-Wasserstein distance $W_p(\mu,\nu)$.
Let $\prob_p(\RR^m)$ be the space of probability densities on $\RR^m$ that have at least the $p$-th moment.
The distance space $\prob_p(\RR^m)$ equipped with $L^p$-Wasserstein distance $W_p$ is called the \emph{$L^p$-Wasserstein space}.
Furthermore, the $L^2$-Wasserstein space $(\spW,W_2)$ admits the \emph{Wasserstein metric} $\metW$,
which is an infinite-dimensional Riemannian metric that induces the $L^2$-Wasserstein distance as the geodesic distance.
Owing to $\metW$, the $L^2$-Wasserstein space is an infinite-dimensional manifold.
On $\spW$, we can introduce the tangent space $T_\mu \spW$ at $\mu \in \spW$, and the gradient operator $\grad$,
which are fundamentals to define \emph{Wasserstein gradient flow}. See \refsec{quick} for more details.

\subsection*{Organization of This Paper}
In \refsec{quick}, we describe the framework of transport analysis, which combines a quick introduction to dynamical systems theory, optimal transport theory, and Wasserstein gradient flow.
In \refsec{sdae} and \ref{sec:ddae}, we specify the transport maps of shallow, deep, and infinitely deep DAEs, and we give their statistical interpretations. In \refsec{examples}, we present analytic examples and the results of numerical experiments.
In \refsec{stack.dae}, we prove the equivalence between the stacked DAE and the composition of DAEs.
In \refsec{intrep.flowrep}, we develop the integral representation of the flow representation.

\subsection*{Remark}
After the initial submission of the manuscript in 2016,
the present manuscript has been substantially reorganized and updated.
The authors presented the digests of some results from \refsec{sdae}, \ref{sec:ddae} and \ref{sec:intrep.flowrep} in two workshops \citep{Sonoda2017,Sonoda2017a}.

\section{Transport Analysis of Deep Neural Networks} \label{sec:quick}

In the transport analysis,
we regard a deep neural network as a transport map,
and we track the flow in three scales: microscopic, mesoscopic, and macroscopic.
Wasserstein geometry provides a unified framework for bridging these three scales.
In each scale, we analyze three profiles of the flow: dynamical system, pushforward measure, and Wasserstein gradient flow.

\begin{figure}[t]
\centering
\begin{tabular}{ccc}
	\begin{minipage}{0.24\textwidth}
		\centering
		\includegraphics[width=\textwidth]{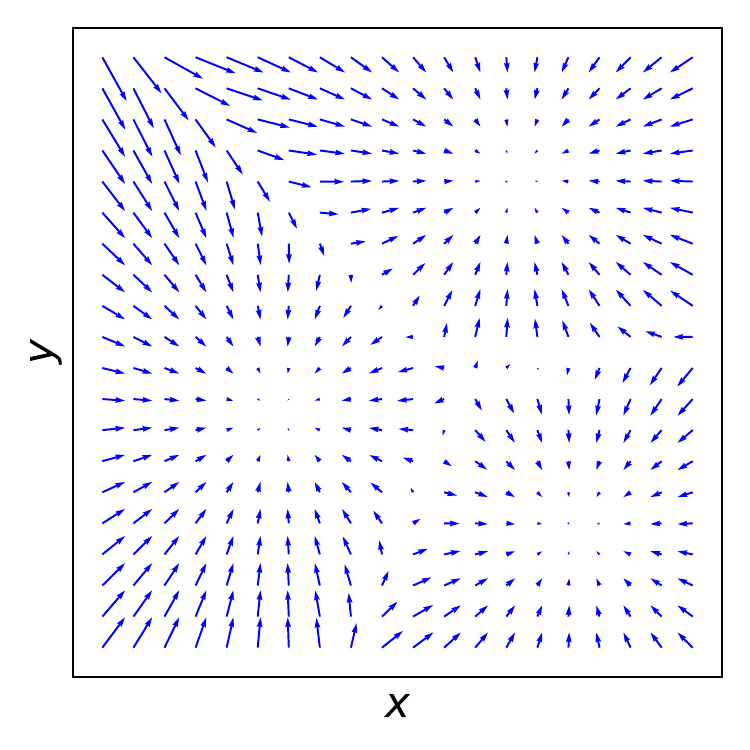}
	\end{minipage}&
	\begin{minipage}{0.33\textwidth}
	\centering
	\includegraphics[width=\textwidth]{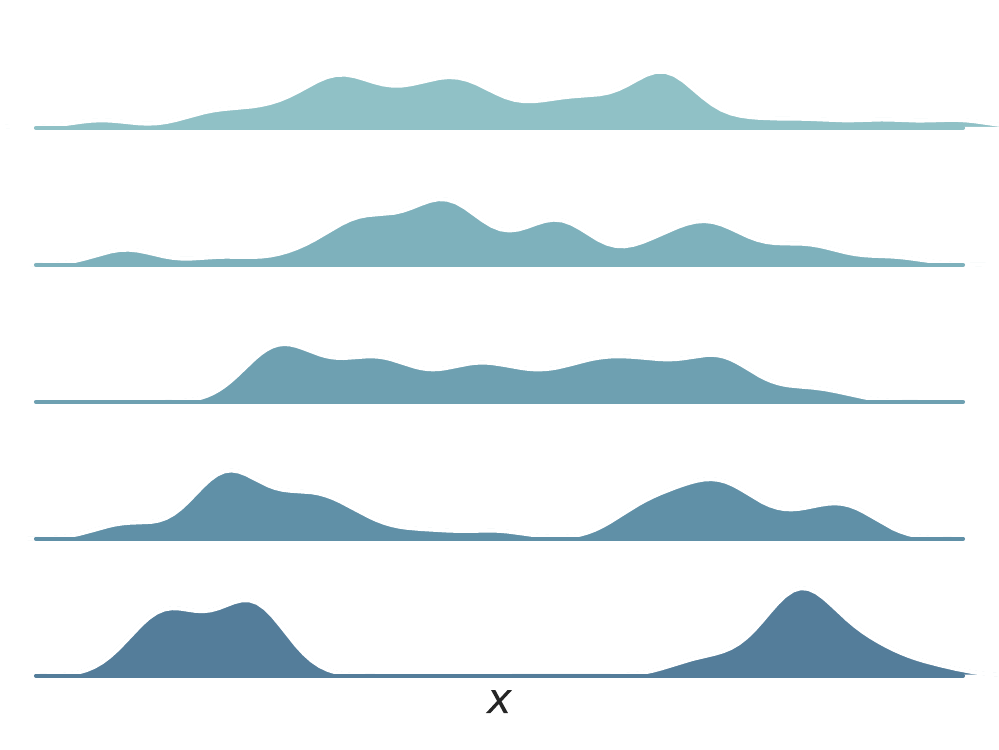}
	\end{minipage}&
	\begin{minipage}{0.33\textwidth}
	\centering
	\includegraphics[width=\textwidth]{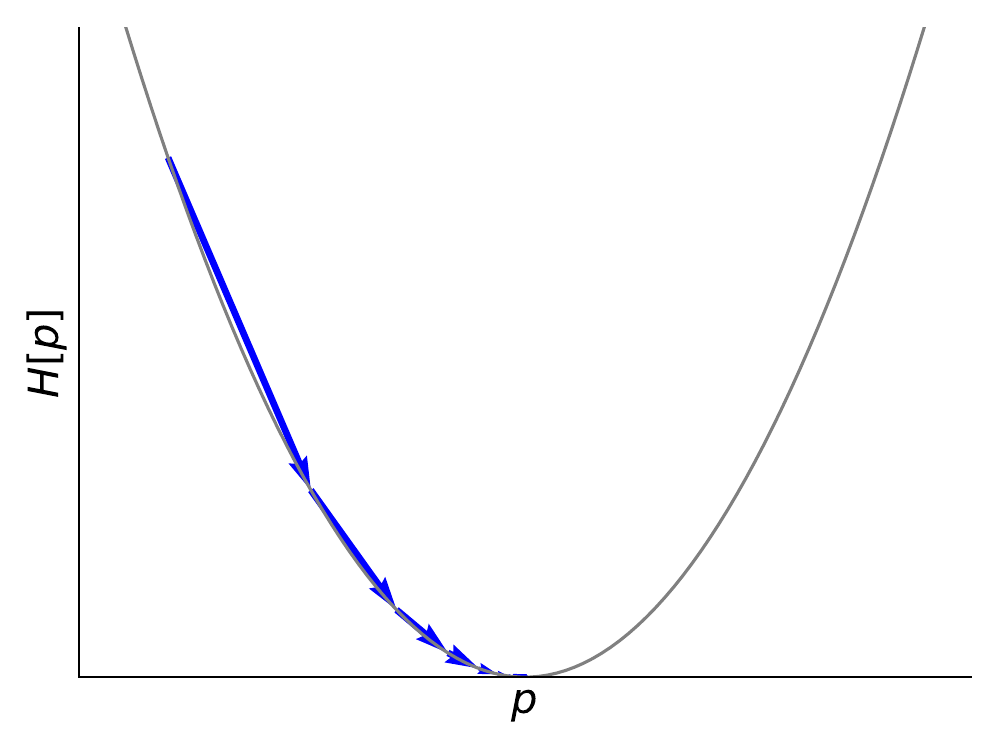}
	\end{minipage}\\
	\end{tabular}
	\caption{Three profiles of a flow analyzed in the transport analysis: dynamical system in $\RR^m$ described by vector field (or transport map)  (\emb{left}), pushforward measure described by continuity equation in $\RR^m$ (\emb{center}), and Wasserstein gradient flow in $\spW(\RR^m)$ (\emb{right}).}
\end{figure}

First, on the microscopic scale, we analyze the transport map $\gg_t : \RR^m \to \RR^m$, which simply describes the transportation of every point.
In continuum mechanics, this viewpoint corresponds to the Eulerian description.
The transport map $\gg_t$ is often associated with a velocity field $\vf_t$ that summarizes all the behavior of $\gg_t$ by an ODE or the continuous dynamical system: $\partial_t \gg_t( \gg_t(\xx) ) = \vf_t( \gg_t(\xx) )$.
We note that, as suggested by chaos theory, it is generally difficult to track a continuous dynamics.

Second, on the mesoscopic scale, we analyze the pushforward $\data_t$ or the time evolution of the data distribution.
In continuum mechanics, this viewpoint corresponds to the Lagrangian description.
When the transport map is associated with a vector field $\vf_t$, then the corresponding distributions evolve according to a partial differential equation (PDE) or the \emph{continuity equation} $\partial_t \data_t = - \nabla \cdot [ \vf_t \data_t ]$.
We note that, as suggested by fluid dynamics, it is generally difficult to track a continuity equation.

Finally, on the macroscopic scale, we analyze the Wasserstein gradient flow or the trajectories of time evolution of $\data_t$ in the space $\prob(\RR^m)$ of probability distributions on $\RR^m$.
When the transport map is associated with a vector field $\vf_t$, then there exists a time-independent potential functional $\free$ on $\prob(\RR^m)$ such that an evolution equation or the Wasserstein gradient flow $\dot{\data_t} = - \grad \free[\data_t]$ coincides with the continuity equation.
We remark that tracking a Wasserstein gradient flow may be easier compared to the two above-mentioned cases, because the potential functional is independent of time.

\subsection{Transport Map and Flow}

In the broadest sense, a \emph{transport map} is simply a measurable map $\gg : M \to N$ between two probability spaces $M$ and $N$ \citep[see Definition~1.2 in][for example]{Villani2009}.
In this study, we use the term as an update rule.
Depending on the context, we distinguish the term ``flow'' from ``transport map.''
While a flow is associated with a continuous dynamical system, a transport map is associated with a discrete dynamical system.
We understand that a transport map arises as a discretization of a flow.
An ordinary DNN coincides with a transport map, and the depth continuum limit coincides with a flow.

\begin{dfn}
A transport map $\gg : \RR^m \to \RR^m$ is a measurable map given by
\begin{align}
\begin{cases}
\dmap_t(\xx) = \xx + \ff_t(\xx), &\xx \in \RR^m, \ t > 0 \\
\dmap_0(\xx) = \xx, &\xx \in \RR^m, \ t=0,
\end{cases}
\label{eq:dmap}
\end{align}
with an update vector $\ff_t$.
\end{dfn}
\begin{dfn}
A flow $\cmap_t$ is given by an ordinary differential equation (ODE),
\begin{align}
\begin{cases}
\dot{\cmap_t}(\xx) = \vf_t(\cmap_t(\xx)), &\xx \in \RR^m, \ t > 0 \\
\cmap_0(\xx) = \xx, &\xx \in \RR^m,\ t=0,
\end{cases} \label{eq:cmap}
\end{align}
with a velocity field $\vf_t$.
\end{dfn}

In particular, we are interested in the case when the update rule \refeq{dmap} is a tangent line approximation of a flow \refeq{cmap}.
i.e., $\gg_t$ satisfies
\begin{align}
\lim_{t \to 0} \frac{\gg_{t}(\xx) - \xx}{t} =  \vf_0( \xx ), \quad \xx \in \RR^m \label{eq:tangent}
\end{align}
for some $\vf_t$.
In this case, the velocity field $\vf_t$ is the only parameter that determines the transport map.


\subsection{Pushforward Measure and Continuity Equation}
In association with the mass transportation $\xx \mapsto \gg_t(\xx)$, the data distribution $\data_0$ itself changes its shape to, say, $\data_t$ (see \reffig{dae1dim}, for example).
Technically speaking, $\data_t$ is called (the density of) the \emph{pushforward measure} of $\data_0$ by $\gg_t$, and it is denoted by $\gg_{t\sharp} \data_0$.

\begin{dfn}
Let $\mu$ be a Borel measure on $M$ and $\gg : M \to N$ be a measurable map.
Then, $\gg_\sharp \mu$ denotes the image measure (or pushforward) of $\mu$ by $\gg$.
It is a measure on $N$, defined by $(\gg_\sharp \mu)(B) = \mu \circ \gg^{-1}(B)$ for every Borel set $B \subset N$.
\end{dfn}

The pushforward $\data_t$ is calculated by the change-of-variables formula.
In particular, the following extended version by \citet[Theorem~3.9]{Evans2015} from geometric measure theory 
is useful.
\begin{fact} \label{fact:cov.singular}
    Let $\gg : \RR^m \to \RR^n$ be Lipschitz continuous, $m \leq n$, and $\data$ be a probability density on $\RR^m$.
    Then, the pushforward $\gg_\sharp \data$ satisfies
    \begin{align}
        \gg_\sharp \data \circ \gg(\xx) [\nabla \gg](\xx) = \data(\xx), \quad \almost \xx. \label{eq:cov.singular}
    \end{align}
    Here, the Jacobian is defined by
    \begin{align}
        [\nabla \gg] = \sqrt{ \det| (\nabla \gg)^* \circ (\nabla \gg)  | }.
    \end{align}
\end{fact}

The continuity equation describes the one-to-one relation between a flow and the pushforward.
\begin{fact} \label{fact:contieq}
Let $\cmap_t$ be the flow of an ODE \refeq{cmap} with vector field $\vf_t$. Then, the pushforward $\data_t$ of the initial distribution $\data_0$ evolves according to the \emph{continuity equation}
\begin{align}
\partial_t \data_t(\xx) = - \nabla \cdot[ \data_t(\xx) \vf_t(\xx) ], \quad \xx \in \RR^m, \, t \geq 0. \label{eq:contieq}
\end{align}
Here, $\nabla \cdot$ denotes the divergence operator in $\RR^m$.
\end{fact}
The continuity equation is also known as the \emph{conservation of mass formula}, and this relation between the partial differential equation (PDE) \refeq{contieq} and the ODE \refeq{cmap} is a well-known fact in continuum physics \citep[pp.19]{Villani2009}.
See \refapp{proof.contieq} for a sketch of the proof and \citet[\S~8]{Ambrosio2008} for more detailed discussions.

\subsection{Wasserstein Gradient Flow Associated with Continuity Equation} \label{sec:wgrad}

In addition to the ODE and PDE in $\RR^m$, we introduce the third profile: the \emph{Wasserstein gradient flow} or the evolution equation in the space of the probability densities on $\RR^m$.
The Wasserstein gradient flow has
a distinct advantage that the potential functional $\free$ of the gradient flow is independent of time $t$; on the other hand, the vector field $\vf_t$ is usually time-dependent.
Furthermore, it often facilitates the understanding of transport maps because we will see that both the Boltzmann entropy and the Renyi entropy are examples of $\free$.

Let $\spW(\RR^m)$ be the $L^2$-Wasserstein space defined in \refsec{quick.intro},
and let $\data_t \in \spW(\RR^m)$ be the solution of the continuity equation \refeq{contieq} with initial distribution $\data_0  \in \spW(\RR^m)$.
Then, the map $t \mapsto \data_t$ plots a curve in $\spW(\RR^m)$.
According to the Otto calculus \citep[\S~23]{Villani2009}, this curve coincides with a functional gradient flow in $\spW(\RR^m)$, called the Wasserstein gradient flow, with respect to some \emph{potential functional} $\free : \spW(\RR^m) \to \RR$.

Specifically,
we further assume that the vector field $\vf_t$ is given by the gradient vector field $\nabla V_t$ of a potential function $V_t : \RR^m \to \RR$.
\begin{fact} \label{fact:wgflow}
Assume that $\data_t$ satisfies the continuity equation with the gradient vector field,
\begin{align}
\partial_t \data_t = - \nabla \cdot[ \data_t \nabla V_t ],
\end{align}
and that we have found $\free$ that satisfies the following equation:
\begin{align}
\frac{\dd }{\dd t} \free[\data_t] = \int_{\RR^m} \potential_t (\xx) [\partial_t \data_t] (\xx) \dd \xx. \label{eq:cond.gflow}
\end{align}
Then, the \emph{Wasserstein gradient flow}
\begin{align}
\frac{\dd}{\dd t} \data_t = - \grad \free[\data_t], \label{eq:gflow}
\end{align}
coincides with the continuous equation.
\end{fact}
Here, $\grad$ denotes the gradient operator on $L^2$-Wasserstein space $\spW(\RR^m)$ explained in \refsec{quick.intro}.
While \refeq{gflow} is an evolution equation or an ODE in $\spW(\RR^m)$,
\refeq{contieq} is a PDE in $\RR^m$. Hence, we use different notations for the time derivatives, $\frac{\dd}{\dd t}$ and $\partial_t$.

\section{Denoising Autoencoder} \label{sec:sdae}

We formulate the denoising autoencoder (DAE) as a variational problem, and we show that the minimizer $\gg^*$ or the training result is a transport map.
Even though the term ``DAE'' refers to a training procedure of neural networks,
we refer to the minimizer of DAE also as a ``DAE.''
We further investigate the initial velocity vector field $\partial_t \gg_{t=0}$ for mass transportation,
and we show that the data distribution $\data_t$ evolves according to the continuity equation.

For the sake of simplicity, we assume that the hidden unit number of NNs is sufficiently large (or infinite), and thus the NNs can always attain the minimum. Furthermore, we assume the the size of data set is sufficiently large (or infinite).
In the case when the hidden unit number and the size of data set are both finite, we understand the DAE $\gg$ is composed of the minimizer $\gg^*$ and the residual term $\hh$. Namely, $\gg = \gg^* + \hh$. However, theoretical investigations on the approximation and estimation error $\hh$ remain as our future work.

\subsection{Training Procedure of DAE}
Let $\xx$ be an $m$-dimensional random vector that is distributed according to the data distribution $\data_0$,
and let $\cx$ be its corruption defined by
\begin{align*}
\cx = \xx + \eps, \quad \eps \sim \noise_t
\end{align*}
where $\noise_t$ denotes the noise distribution parametrized by variance $t \geq 0$.
A basic example of $\noise_t$ is the Gaussian noise with mean $0$ and variance $t$, i.e., $\noise_t = N(0,tI)$.

The DAE is a function that is trained to remove corruption $\cx$ and restore it to the original $\xx$;
this is equivalent to finding a function $\gg$ that minimizes an objective function, i.e.,
\begin{align}
L[\gg] := \EE_{\xx,\cx} | \gg( \cx ) - \xx |^2. \label{eq:objective}
\end{align}
Note that as long as
$\gg$
is a universal approximator and can thus attain the minimum, it need not be a neural network.
Specifically, our analysis in this section and the next section is applicable to a wide range of learning machines.
Typical examples of $\gg$ include neural networks with a sufficiently large number of hidden units, splines \citep{Wahba1990}, kernel machines \citep{Shawe-Taylor2004} and ensemble models \citep{Schapire2012}.

\subsection{Transport Map of DAE}

\begin{thm}\label{thm:sonoda}
    \citep[Modification of Theorem~1 by][]{Alain2014}.
	The global minimum $\gg^*_t$ of $L[\gg]$ is attained at
\begin{align}
\gg_t^*(\cx)
&= \frac{1}{\noise_t * \data_0(\cx)} \int_{\RR^m} \xx \noise_t(\cx-\xx) \data_0(\xx) \dd \xx, \label{eq:alain} \\
&= \cx \underbrace{- \frac{1}{\noise_t * \data_0(\cx)} \int_{\RR^m} \eps \noise_t(\eps) \data_0 (\cx - \eps) \dd \eps}_{=: \, \ff_t (\cx)},
\label{eq:sonoda}
\end{align}
where $*$ denotes the convolution operator.
\end{thm}
Here, the second equation is simply derived by changing the variable $\xx \gets \cx - \eps$
(see \refapp{proof.sonoda} for the complete proof, where we used the calculus of variations).
Note that this calculation first appeared in
\citet[Theorem~1]{Alain2014}, where the authors obtained \refeq{alain}.

The DAE $\gg^*_t(\xx)$ is composed of the identity term $\xx$ and the denoising term $\ff_t(\xx)$.
If we assume that $\noise_t \to \delta_t$ as $t \to 0$,
then in the limit $t \to 0$, the denoising term $\ff_t(\xx)$ vanishes and DAE reduces to a traditional autoencoder.
We reinterpret the DAE $\gg^*_t(x)$ as a \emph{transport map with transport time $t$} that transports the mass at $\xx \in \RR^m$ toward $\xx + \ff_t(\xx) \in \RR^m$ with displacement vector $\ff_t(\xx)$.

\subsection{Statistical Interpretation of DAE} \label{sec:stat.dae}
In statistics, \refeq{sonoda} is known as Brown's representation of the posterior mean \citep{George2006}.
This is not just a coincidence, because the DAE $\gg^*_t$ is an estimator of the mean.
Recall that a DAE is trained to retain the original vector $\xx$, given its corruption $\cx = \xx + \eps$.
At least in principle, this is nonsense because to retain $\xx$ from $\cx$ means to reverse the random walk $\cx = \xx + \eps$
(in \reffig{dae1dim}, the multimodal distributions $\data_{0.5}$ and $\data_{1.0}$ indicate its difficulty).
Obviously, this is an inverse problem or a statistical estimation problem of the latent vector $\xx$, given the noised observation $\cx$ with the observation model $\cx = \xx + \eps$.
According to a fundamental fact of estimation theory, the minimum mean squared error (MMSE) estimator of $\xx$ given $\cx$ is given by the posterior mean $\EE[\xx | \cx]$. In our case, the posterior mean equals $\gg^*_t$.
\begin{align}
\EE[\xx | \cx]
= \frac{\int_{\RR^m} \xx p(\cx \mid \xx) p(\xx) \dd \xx}{\int_{\RR^m} p(\cx \mid \xx') p(\xx') \dd \xx'} = \frac{1}{\noise_t * \data_0(\cx)} \int_{\RR^m} \xx \noise_t(\cx-\xx) \data_0(\xx) \dd \xx = \gg^*_t(\cx).
\end{align}
Similarly, we can interpret the denoising term $\ff_t(\cx)$ as the posterior mean $\EE[\eps | \cx]$ of noise $\eps$ given observation $\cx$.

\subsection{Examples: Gaussian DAE}

When the noise distribution is Gaussian with mean $0$ and covariance $tI$, i.e.,
\begin{align*}
\noise_t(\eps) = \frac{1}{(2 \pi t)^{m/2}} e^{ -|\eps|^2/2t },
\end{align*}
the transport map is calculated as follows.
\begin{thm} The transport map $\gg^*_t$ of Gaussian DAE is given by
	\begin{align}
	\gg^*_t(\cx) = \cx + t \nabla \log[ \noise_t * \data_0](\cx).  \label{eq:gdae.sonoda}
	\end{align}
\end{thm}

\proofhere
	The proof is straightforward by using Stein's identity,
	\begin{align*}
	-t \nabla \noise_t(\eps) = \eps \, \noise_t(\eps),
	\end{align*}
	which is known to hold only for Gaussians.
\begin{align}
\gg^*_t(\cx)
&= \cx - \frac{1}{\noise_t * \data_0(\cx)} \int_{\RR^m} \eps \noise_t(\eps) \data_0 (\cx - \eps) \dd \eps \nonumber \\
&= \cx +\frac{1}{\noise_t * \data_0(\cx)} \int_{\RR^m} t \nabla \noise_t(\eps) \data_0 (\cx - \eps) \dd \eps \nonumber \\
&= \cx + \frac{t \nabla \noise_t * \data_0(\cx)}{\noise_t * \data_0(\cx)} \nonumber \\
&= \cx +t \nabla \log [ \noise_t * \data_0 (\cx)]. \qedhere
\end{align}

\begin{thm} \label{thm:dae.pfinit}
	At the initial moment $t \to 0$, the pushforward $\data_t$ of Gaussian DAE satisfies the \emph{backward heat equation}
	\begin{align}
	\partial_t \data_{t=0}(\xx) = -\triangle \data_0(\xx), \quad \xx \in \RR^m, \label{eq:dae.pf.init}
	\end{align}
	where $\triangle$ denotes the Laplacian.
\end{thm}
\proofhere
The initial velocity vector is given by the \emph{Fisher score}
\begin{align}
\partial_t \gg^*_{t=0} (\xx) = \lim_{t \to 0} \frac{\gg^*_t(\xx) - \xx}{t} = \nabla \log \data_0 (\xx). \label{eq:dae.tp.init}
\end{align}
Hence, by substituting the score \refeq{dae.tp.init} in the continuity equation \refeq{contieq}, we have
\begin{align*}
\partial_t \data_{t=0}(\xx)
= -\nabla \cdot [ \data_{0} (\xx) \nabla \log \data_0(\xx) ]
= -\nabla \cdot [ \nabla \data_0(\xx) ]
= -\triangle \data_0(\xx). \qedhere
\end{align*}

\begin{figure}[t]
	\centering
	\includegraphics[width=0.5\textwidth]{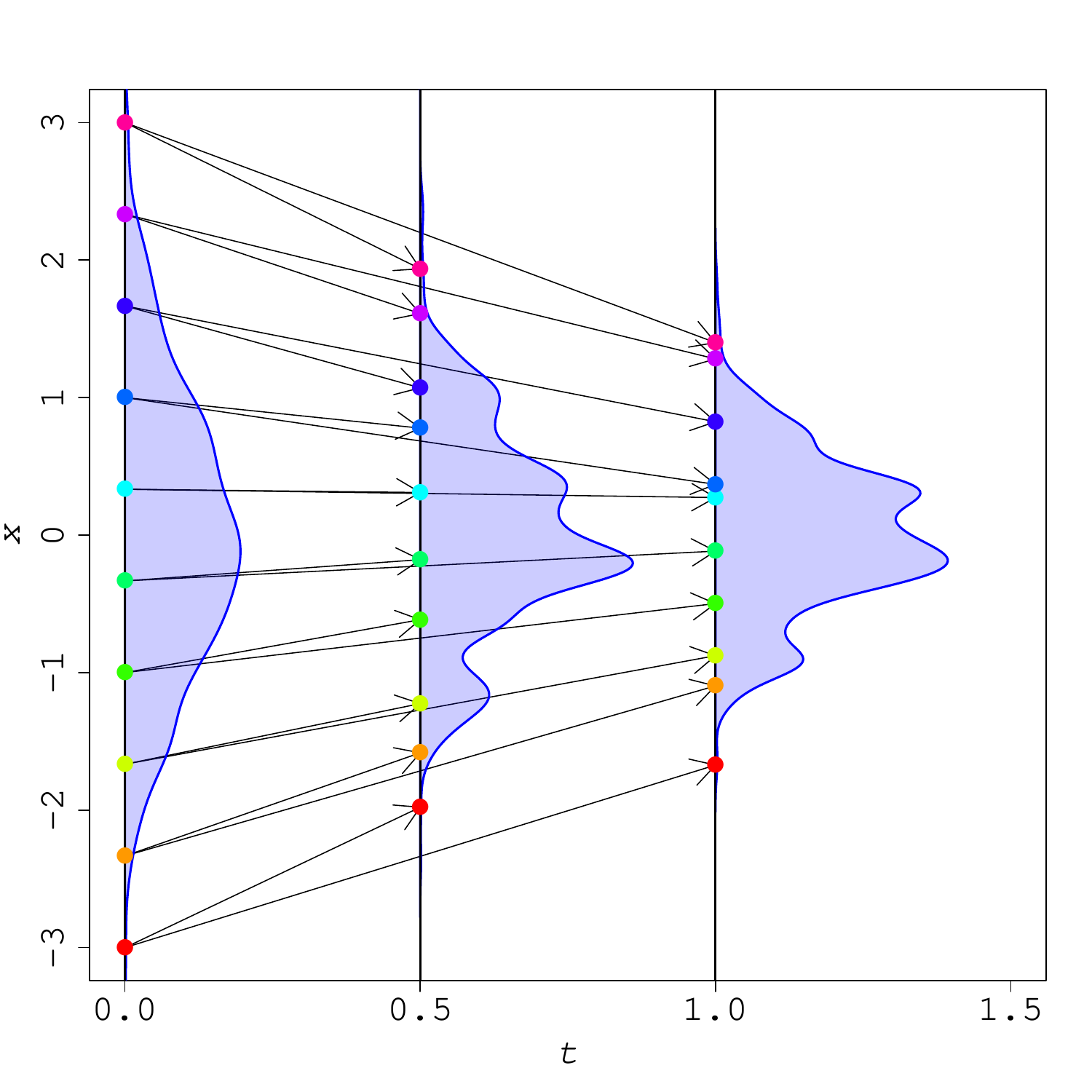}
	\caption{Shallow Gaussian DAE, which is one of the most fundamental versions of DNNs, transports mass, from the left to the right, to decrease the Shannon entropy of data. The $x$-axis represents the $1$-dimensional input/output space, the $t$-axis represents the variance of the Gaussian noise, and $t$ is the transport time. The leftmost distribution depicts the original data distribution $\data_0 = N(0,1)$. The middle and rightmost distributions depict the pushforward $\data_t = \gg_{t\sharp} \data_0$, associated with the transportation by two DAEs with noise variance $t=0.5$ and $t=1.0$, respectively. As $t$ increases, the variance of the pushforward decreases.}
	\label{fig:dae1dim}
\end{figure}

The backward heat equation (BHE) rarely appears in nature. However, of course, the present result is not an error.
As mentioned in \refsec{stat.dae}, the DAE solves an estimation problem. Therefore, in the sense of the mean, the DAE behaves as time reversal.
We remark that, as shown by \reffig{dae1dim}, a training result of a DAE with a real NN on a finite data set does not converge to a perfect time reversal of a diffusion process.

\clearpage

\section{Deep DAEs} \label{sec:ddae}
We introduce the composition $\gg_L \circ \cdots \circ \gg_0$ of DAEs $\gg_\ell : \RR^m \to \RR^m$
and its continuum limit: the continuous DAE $\cdae_t : \RR^m \to \RR^m$.
We can understand the composition of DAEs as the \emph{Euler scheme} or the \emph{broken line approximation} of a continuous DAE.

For the sake of simplicity, we assume that the hidden unit number of NNs is infinite, and that the size of data set is infinite.

\subsection{Composition of DAEs} \label{sec:compdae}
We write $0 = t_0 < t_1 < \cdots < t_{L+1} = t$.
We assume that the input vector $\xx_0 \in \RR^m$ is subject to a data distribution $\data_0$.
Let $\gg_0 : \RR^m \to \RR^m$ be a DAE that is trained on $\data_0$ with noise variance $t_1 - t_0$.
Then, let $\xx_1 := \gg_0(\xx_0)$, which is a random vector in $\RR^m$ that is subject to the pushforward $\data_1 := \gg_{0 \sharp} \data_0$.
We train another DAE $\gg_1 : \RR^m \to \RR^m$ on $\data_1$ with noise variance $t_2- t_1$.
By repeating the procedure, we obtain $\gg_{\ell}(\xx_{\ell})$ from $\xx_{\ell-1}$ that is subject to $\data_{\ell} := \gg_{(\ell-1) \sharp} \data_{\ell-1}$. 

For the sake of generality, we assume that each component DAE is given by
\begin{align}
&\gg_\ell(\xx) = \xx + (t_{\ell+1} - t_{\ell}) \nabla V_{t_\ell}(\xx), \quad (\ell = 0, \ldots, L)
\end{align}
where $V_{t_\ell}$ denotes a certain potential function.
For example, the Gaussian DAE satisfies the requirement because $V_{t_\ell} = \log [\noise_{t_\ell} * \data_{t_\ell}]$.

We abbreviate the composition of DAEs by
\begin{align}
\gg_{0:L}^t(\xx) := \gg_L \circ \cdots \circ \gg_0(\xx).
\end{align}
By definition, the ``velocity'' of a composition of DAEs coincides with the vector field
\begin{align}
\frac{\gg_{0:\ell}^{t_{\ell+1}}(\xx) - \gg_{0:(\ell - 1)}^{t_{\ell}}(\xx)}{t_{\ell + 1}-t_\ell} = \nabla V_{t_\ell}(\xx).
\end{align}
\begin{figure}[t]
	\centering
	\includegraphics[width=\textwidth]{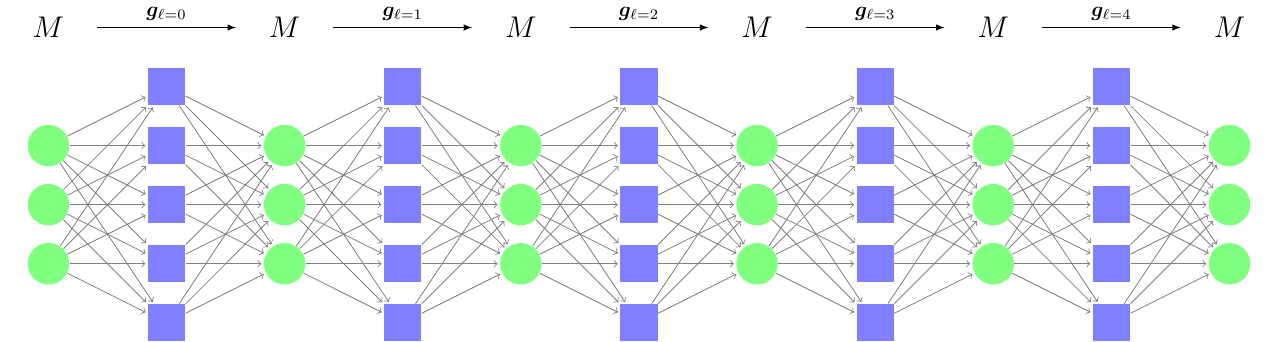}
	\caption{Composition of DAEs $\gg_{0:4}^t : M \to M$, or the composite of five shallow DAEs $M \to M$, where $M = \RR^3$}
\end{figure}

\subsection{Continuous DAE}
We fix the total time $t$, take the limit $L \to \infty$ of the layer number $L$, and introduce the continuous DAE as the limit of the ``infinite composition of DAEs'' $\lim_{L \to \infty} \gg_{0:L}^t$.
\begin{dfn}
	We call the solution operator or flow $\cdae_t : \RR^m \to \RR^m$ of the following dynamical systems as the \emph{continuous DAE} associated with vector field $\nabla V_t$.
	\begin{align}
	\frac{\dd }{\dd t} \xx(t) = \nabla V_t( \xx(t) ), \quad t \geq 0. \label{eq:dfn.cdae}
	\end{align}
\end{dfn}
\begin{proof}
According to the Cauchy-Lipschitz theorem or the Picard-Lindel\"{o}f theorem,
when the vector field $\nabla V_t$ is continuous in $t$ and Lipschitz in $\xx$,
the limit $\lim_{L \to \infty} \gg_{0:L}$ converges to a continuous DAE \refeq{dfn.cdae}
because the trajectory $t \mapsto \gg_{0:L}(x_0)$ corresponds to a broken line approximation of the integral curve $t \mapsto \cdae_t(x)$.
\end{proof}

The following properties are immediate from \reffact{contieq} and \reffact{wgflow}.
Let $\cdae_t : \RR^m \to \RR^m$ be the continuous DAE associated with vector field $\nabla V_t$. Given the data distribution $\data_0$, the pushforward $\data_t := (\cdae_t)_\sharp \data_0$ evolves according to the continuity equation
	\begin{align}
	    \partial_t \data_t(\xx) = - \nabla \cdot [ \data_t(\xx) \nabla V_t(\xx) ], \quad t \geq 0 \label{eq:conti}
	\end{align}
	and the Wasserstein gradient flow
	\begin{align}
	    \frac{\dd}{\dd t} \data_t = - \grad \free[\data_t], \quad t \geq 0
	\end{align}
	where $\free$ is given by \refeq{cond.gflow}.

\subsection{Example: Gaussian DAE}
We consider a continuous Gaussian DAE $\cdae_t$ trained on $\data_0 \in \spW(\RR^m)$. Specifically, it satisfies
\begin{align}
    \frac{\dd}{\dd t} \xx(t) = \nabla \log [\data_t( \xx(t) )], \quad t \geq 0
\end{align}
with $\data_t := \cdae_{t \sharp} \data_0$.

\begin{thm} \label{thm:cdae.backward}
	The pushforward $\data_t := \cdae_{t \sharp} \data_0$ of the continuous Gaussian DAE $\cdae_t$ is the solution to the initial value problem of the backward heat equation (BHE)
	\begin{align}
	\partial_t \data_t(\xx) = - \triangle \data_t(\xx), \quad \data_{t=0}(\xx) =\data_0(\xx). \label{eq:backward}
	\end{align}
\end{thm}
The proof is immediate from \refthm{dae.pfinit}.

As mentioned after \refthm{dae.pfinit}, the BHE appears because the DAE solves an estimation problem.
We remark that the BHE is equivalent to the following \emph{final value problem} for the ordinary heat equation:
\begin{align*}
\partial_t u_t(\xx) = \triangle u_t(\xx), \quad u_{t = \tbound}(\xx) = \data_0(\xx) \quad \mbox{ for some } \tbound
\end{align*}
where $u_t$ denotes a probability measure on $\RR^m$.
Indeed, $\data_t(\xx) = u_{\tbound - t}(\xx)$ solves \refeq{backward}.
In other words, the backward heat equation describes the time reversal of an ordinary diffusion process.

According to Wasserstein geometry, an ordinary heat equation corresponds to a Wasserstein gradient flow that \emph{increases} the Shannon entropy functional $\ent[\data] := -\int \data(\xx) \log \data(\xx) \dd \xx$ \citep[Th.~23.19]{Villani2009}.
Consequently, we can conclude that the continuous Gaussian DAE is a transport map that \emph{decreases} the Shannon entropy of the data distribution.
\begin{thm}
The pushforward $\data_t := \cdae_{t \sharp} \data_0$  evolves according to the Wasserstein gradient flow with respect to the Shannon entropy
\begin{align}
\frac{\dd}{\dd t} \data_t = - \grad \ent[\data_t], \quad \data_{t=0} = \data_0.
\end{align}
\end{thm}
\begin{proof}
When $\free=\ent$, then $V_t = -\log \data_t$; thus,
\begin{align*}
\grad \ent[\data_t]
= \nabla \cdot [ \data_t \nabla \log \data_t ]
= \nabla \cdot \left[ \nabla \data_t \right]
= \triangle \data_t,
\end{align*}
which means that the continuity equation reduces to the backward heat equation.
\end{proof}

\subsection{Example: Renyi Entropy}
Similarly, when $\free$ is the Renyi entropy
\begin{align*}
\ent^\alpha[\data] := \int_{\RR^m} \frac{\data^\alpha(\xx) - \data(\xx)}{\alpha - 1} \dd \xx,
\end{align*}
then $\grad \ent^\alpha [\data_t] = \triangle \data_t^\alpha$
\citep[see Ex.~15.6 in][for the proof]{Villani2009} and thus the continuity equation reduces to the \emph{backward porous medium equation}
\begin{align}
\partial_t \data_t(\xx) = - \triangle \data_t^\alpha(\xx).
\end{align}

\section{Further Investigations on Shallow and Deep DAEs through Examples} \label{sec:examples}

\subsection{Analytic Examples}
We list analytic examples of shallow and continuous DAEs (see \refapp{anal.examples} for further details, including proofs).
In all the settings,
the continuous DAEs attain a singular measure at some finite $t>0$ with various singular supports that reflect the initial data distribution $\data_0$,
while the shallow DAEs accept any $t>0$ and degenerate to a point mass as $t \to \infty$.

\subsubsection{Univariate Normal Distribution}
When the data distribution is a univariate normal distribution $N(m_0, \sigma_0)$,
the transport map and pushforward for the \emph{shallow} DAE are given by
\begin{align}
g_t(x) &= \frac{\sigma_0^2}{\sigma_0^2 + t}x + \frac{t}{\sigma_0^2 + t} m_0, \label{eq:eg.ust}\\
\data_t &= N\left( m_0, \frac{\sigma_0^2}{(1 + t /\sigma_0^2)^2} \right),\label{eq:eg.usp}
\end{align}
and those of the \emph{continuous} DAE are given by
\begin{align}
g_t(x) &=
\sqrt{ 1 - 2t / \sigma_0^2} (x-m_0) + m_0,\label{eq:eg.uct}\\
\data_t &=
N( m_0, \sigma_0^2 - 2 t).\label{eq:eg.ucp}
\end{align}

\subsubsection{Multivariate Normal Distribution}
When the data distribution is a multivariate normal distribution $N(\mm_0, \Sigma_0)$,
the transport map and pushforward for the \emph{shallow} DAE are given by
\begin{align}
\gg_t(\xx) &=(I + t \Sigma_0^{-1})^{-1}\xx + (I + t^{-1}\Sigma_0)^{-1} \mm_0,\label{eq:eg.mst}\\
\data_t &=N( \mm_0, \Sigma_0(I + t \Sigma_0^{-1})^{-2} ),\label{eq:eg.msp}
\end{align}
and those of the \emph{continuous} DAE are given by
\begin{align}
\gg_t(\xx) &= \sqrt{ I - 2 t \Sigma_0^{-1 }}(\xx - \mm_0) + \mm_0,\label{eq:eg.mct}\\
\data_t &= N( \mm_0, \Sigma_0 - 2 t I ).\label{eq:eg.mcp}
\end{align}

\subsubsection{Mixture of Multivariate Normal Distributions} \label{sec:eg.x}
When the data distribution is a mixture of multivariate normal distributions $\sum_{k=1}^K w_k N(\mm_k, \Sigma_k)$
with the assumption that it is \emph{well separated},
the transport map and pushforward for the \emph{shallow} DAE are given by
\begin{align}
\gg_t(\xx) &=\sum_{k=1}^K \gamma_{kt}(\xx) \left\{ (I+ t \Sigma_k^{-1})^{-1}\xx + (I+t^{-1}\Sigma_k)^{-1}\mm_k \right\},\label{eq:eg.xst}\\
\data_t &\approx \sum_{k=1}^K w_k N( \mm_k, \Sigma_k(I + t \Sigma_k^{-1})^{-2} ),\label{eq:eg.xsp}
\end{align}
with responsibility function
\begin{align}
\gamma_{kt}(\xx) &:= \frac{w_k N(\xx;\mm_k , \Sigma_k + t I)}{\sum_{k=1}^K w_k N(\xx;\mm_k , \Sigma_k + t I)}, \label{eq:eg.xsr} 
\end{align}
and those of the \emph{continuous} DAE are given by
\begin{align}
\gg_t(\xx) &\approx \sqrt{ I - 2 t \Sigma_k^{-1} }( \xx - \mm_k ) + \mm_k,\label{eq:eg.xct}\\
\data_t &=\sum_{k=1}^K w_k N( \mm_k, \Sigma_k - 2 t I ),\label{eq:eg.xcp}
\end{align}
with responsibility function
\begin{align}
\gamma_{kt}(\xx) &:= \frac{w_k N(\xx;\mm_k , \Sigma_k - 2 t I)}{\sum_{k=1}^K w_k N(\xx;\mm_k , \Sigma_k - 2 t I)}. \label{eq:eg.xcr} 
\end{align}
Here, we say that the mixture $\sum_{k=1}^K w_k N(\mm_k, \Sigma_k)$ is well separated when
for every cluster center $\mm_k$, there exists a neighborhood $\Omega_k$ of $\mm_k$ such that $N( \Omega_k ; \mm_k, \Sigma_k ) \approx 1$ and $\gamma_{kt} \approx \ind_{\Omega_k}$.

\subsection{Numerical Example of Trajectories}
We employed $2$-dimensional examples, in order to visualize the difference of vector fields between the shallow and deep DAEs.
In the examples below, every trajectories are drawn into attractors, however the shape of the attractors and the speed of trajectories are significantly different between shallow and deep.

\subsubsection{Bivariate Normal Distribution}

\begin{figure}[t]
	\centering
	\begin{tabular}{cc}
		\begin{minipage}{0.5\hsize}
			\centering
			Conti DAE $\cdae_t$ \\ \vspace{-.5cm}
			\includegraphics[width=\textwidth]{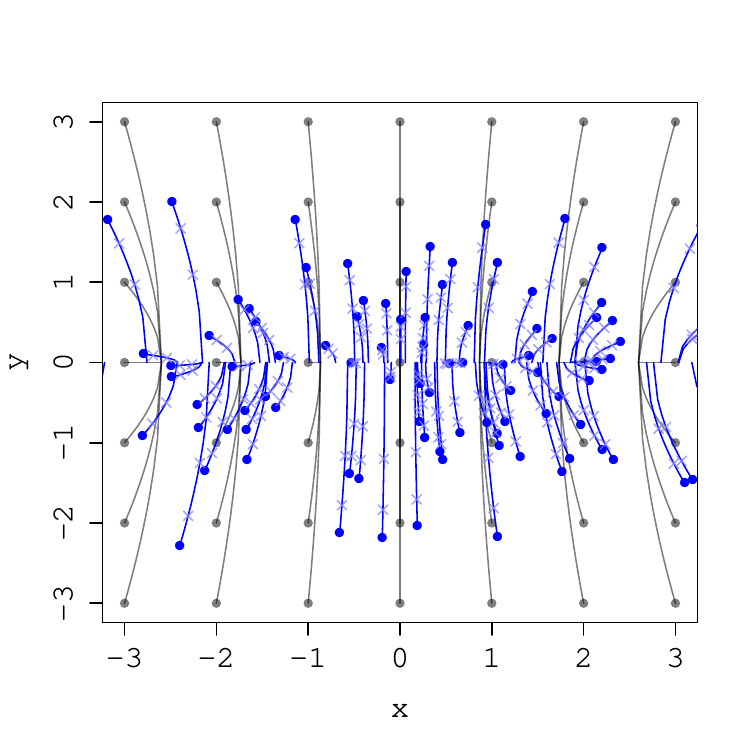}
		\end{minipage} &
		\begin{minipage}{0.5\hsize}
			\centering
			Shallow DAE $\gg_t$ \\ \vspace{-.5cm}
			\includegraphics[width=\textwidth]{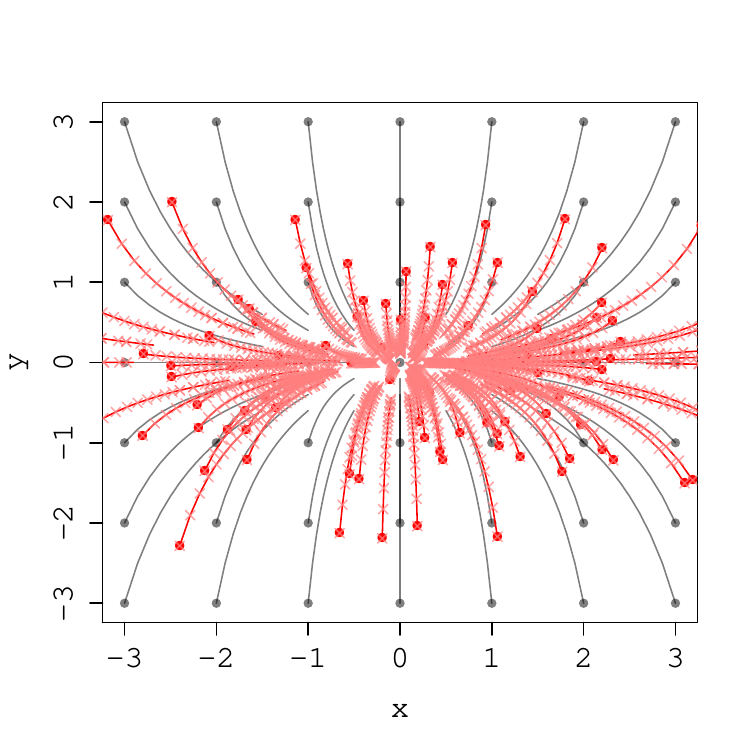}
		\end{minipage} \\
		\begin{minipage}{0.5\hsize}
			\centering
			Comp. DAE $\gg_{0:L}^t \, (\dt=0.05)$ \\ \vspace{-.5cm}
			\includegraphics[width=\textwidth]{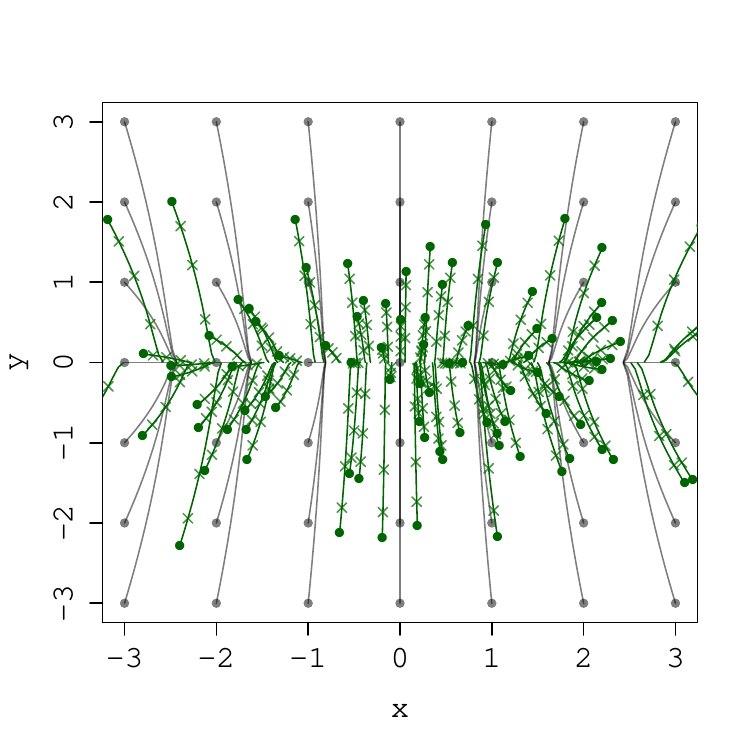}
		\end{minipage} &
		\begin{minipage}{0.5\hsize}
			\centering
			Comp. DAE $\gg_{0:L}^t \, (\dt=0.5)$ \\ \vspace{-.5cm}
			\includegraphics[width=\textwidth]{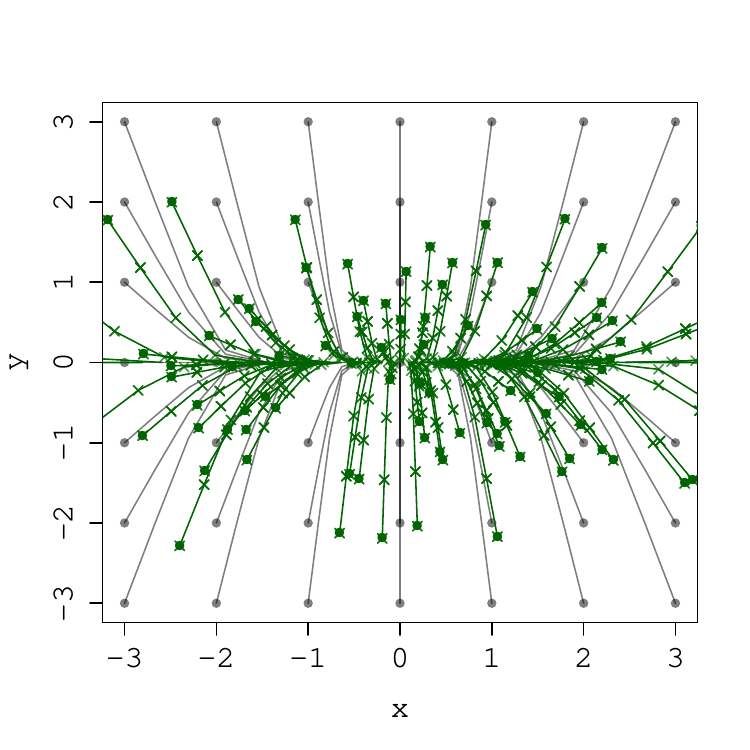}
		\end{minipage} \\
	\end{tabular}
	\caption{
		Trajectories of DAEs trained on the common data distribution \refeq{bivariate} ($\data_0 = N([0,0],\diag[2,1])$).
		The \emb{gray lines} start from the regular grid. The \emb{colored lines} start from the samples drawn from $\data_0$.
		The \emb{midpoints} are plotted every $\dt = 0.2$. Every lines are drawn into attractors.}
	\label{fig:dae.cdae.comdae}
\end{figure}

\reffig{dae.cdae.comdae} compares the trajectories of four DAEs trained on the common data distribution
\begin{align}
\data_0 =
N \left( [0,0], \begin{bmatrix}
2 & 0 \\ 0 & 1
\end{bmatrix} \right). \label{eq:bivariate}
\end{align}
The transport maps for computing the trajectories are given by \refeq{eg.mst} for the shallow DAE and composition of DAEs, and by \refeq{eg.mct} for the continuous DAE. Here, we applied \refeq{eg.mst} multiple times for the composition of DAEs.

The continuous DAE converges to an attractor lying on the $x$-axis at $t = 1/2$.
By contrast, the shallow DAE slows down as $t \to \infty$ and never attains the singularity in finite time.
As $L$ tends to infinity, $\gg_{0:L}^t$ plots a trajectory similar to that of the continuous DAE $\cdae_t$; the curvature of the trajectory changes according to $\dt$.

\subsubsection{Mixture of Bivariate Normal Distributions}
\begin{figure}[t]
	\centering
	\begin{tabular}{cc}
		\begin{minipage}{0.5\hsize}
			\centering
			Conti DAE $\cdae_t$ \\ \vspace{-.5cm}
						\includegraphics[width=\textwidth]{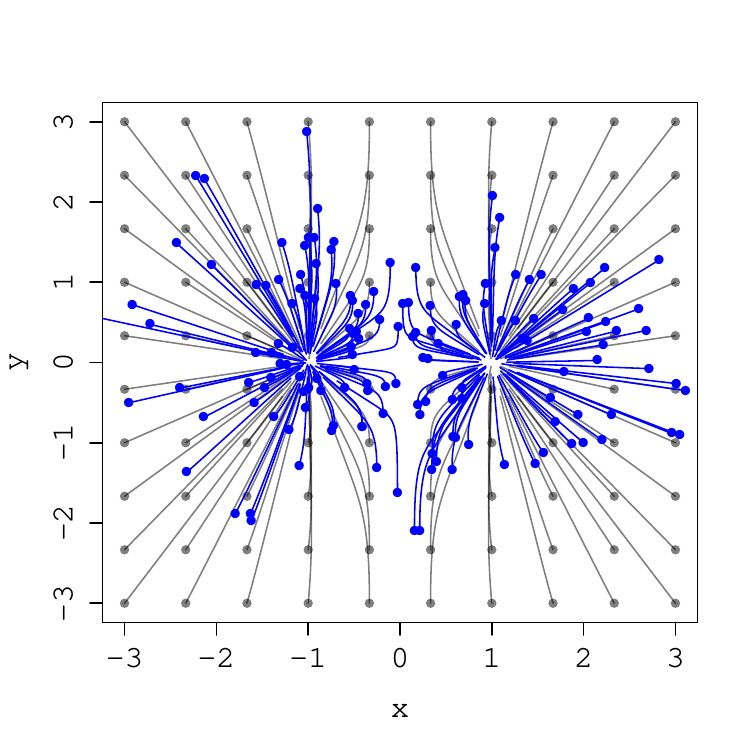}
		\end{minipage} &
		\begin{minipage}{0.5\hsize}
			\centering
			Shallow DAE $\gg_t$ \\ \vspace{-.5cm}
			\includegraphics[width=\textwidth]{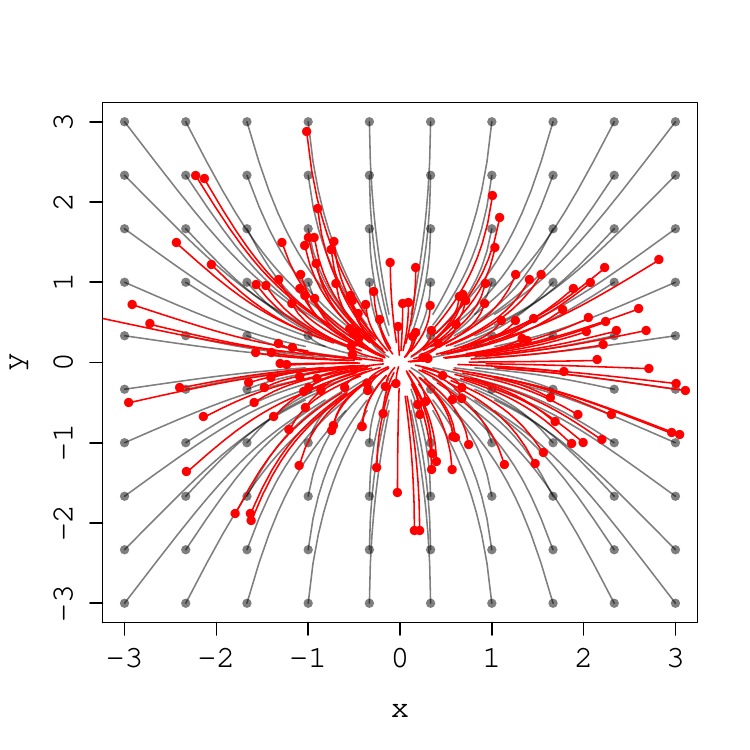}
		\end{minipage} \\
		\begin{minipage}{0.5\hsize}
			\centering
			Comp. DAE $\gg_{0:L}^t \, (\dt=0.05)$ \\ \vspace{-.5cm}
			\includegraphics[width=\textwidth]{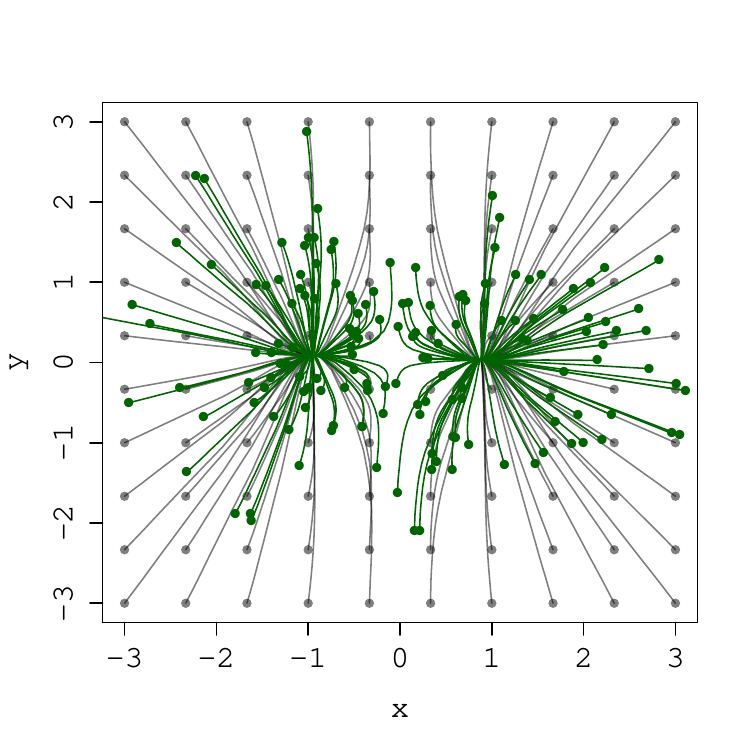}
		\end{minipage} &
		\begin{minipage}{0.5\hsize}
			\centering
			Comp. DAE $\gg_{0:L}^t \, (\dt=0.5)$ \\ \vspace{-.5cm}
\includegraphics[width=\textwidth]{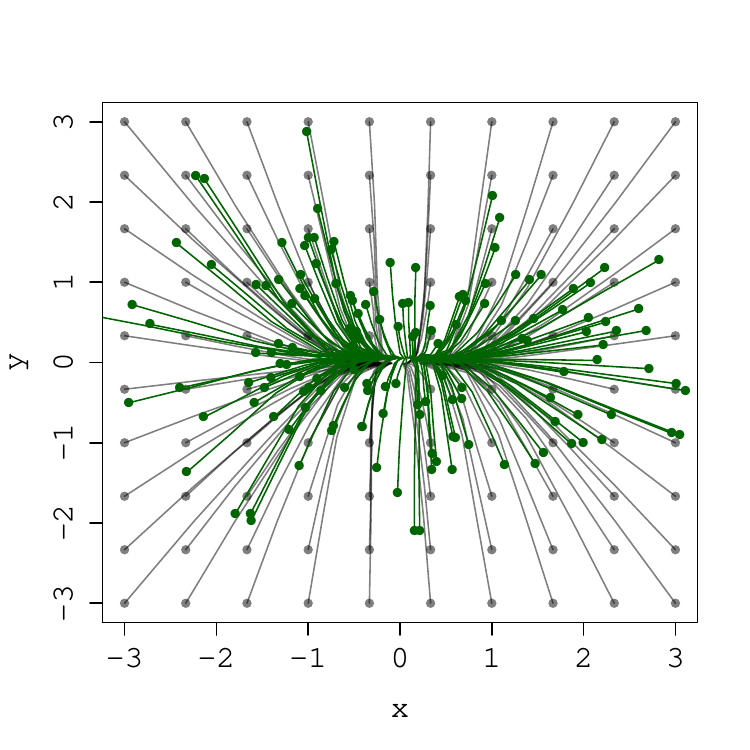}
		\end{minipage} \\
	\end{tabular}
		\caption{Trajectories of DAEs trained on the common data distribution \refeq{gmm1} (a GMM with uniform weight and covariance).
		The \emb{gray lines} start from the regular grid. The \emb{colored lines} start from the samples drawn from $\data_0$.  Every lines are drawn into attractors.}
	  \label{fig:flow.gmm1}
\end{figure}

\begin{figure}[t]
	\centering
	\begin{tabular}{cc}
		\begin{minipage}{0.5\hsize}
			\centering
			Conti. DAE $\cdae_t$ \\ \vspace{-.5cm}
	\includegraphics[width=\textwidth]{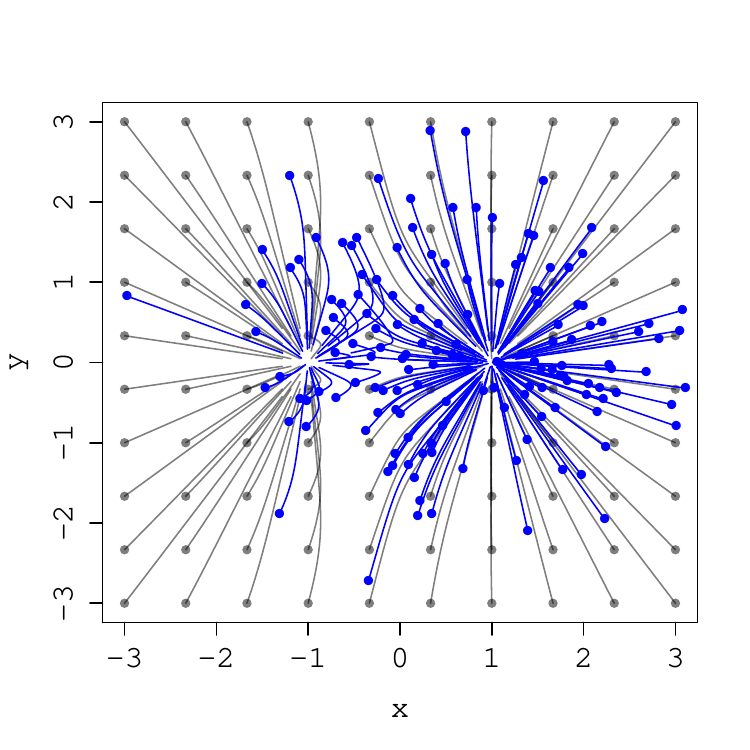}
		\end{minipage} &
		\begin{minipage}{0.5\hsize}
			\centering
			Shallow DAE $\gg_t$ \\ \vspace{-.5cm}
	\includegraphics[width=\textwidth]{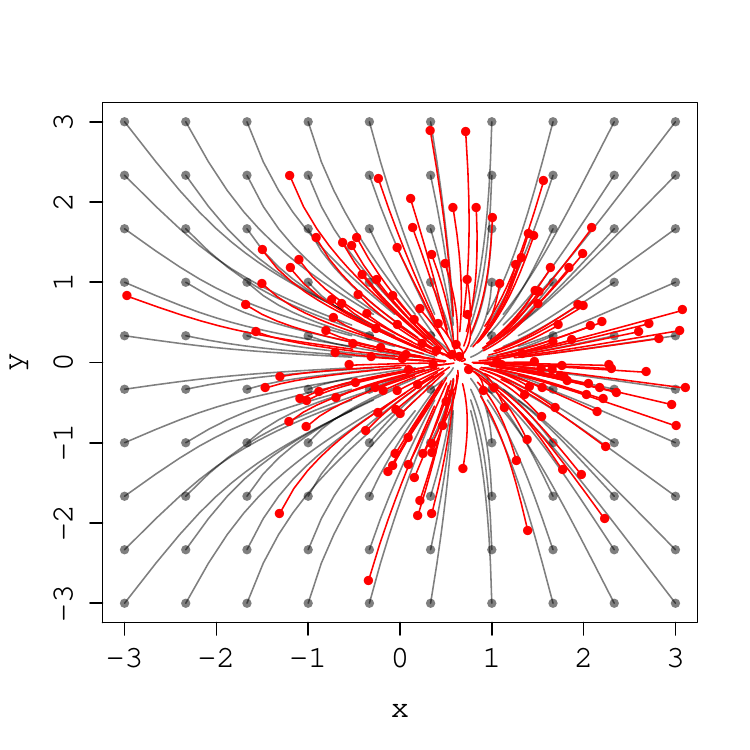}
		\end{minipage} \\
		\begin{minipage}{0.5\hsize}
			\centering
			Comp. DAE $\gg_{0:L}^t \, (\dt=0.05)$ \\ \vspace{-.5cm}
	\includegraphics[width=\textwidth]{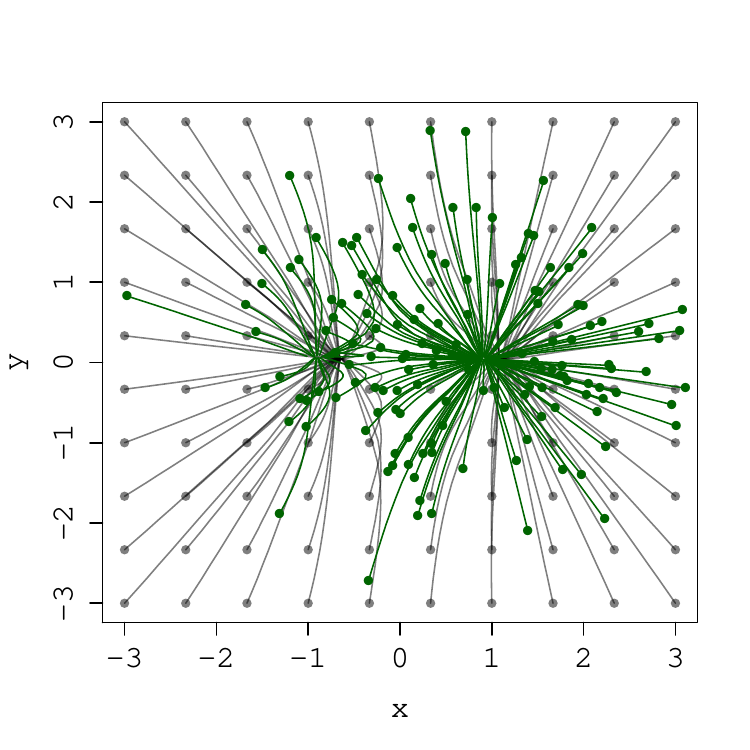}
		\end{minipage} &
		\begin{minipage}{0.5\hsize}
			\centering
			Comp. DAE $\gg_{0:L}^t \, (\dt=0.5)$ \\ \vspace{-.5cm}
	\includegraphics[width=\textwidth]{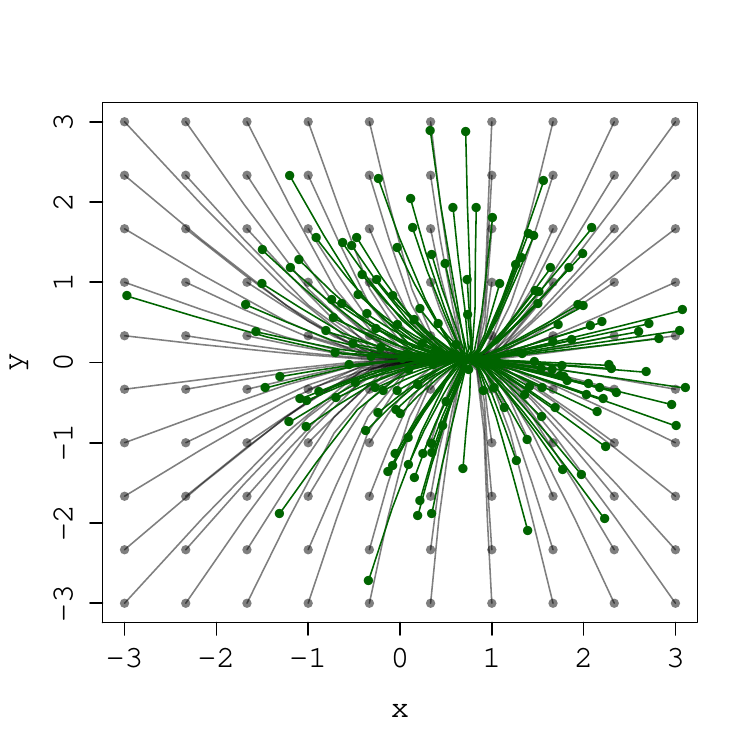}
		\end{minipage} \\
	\end{tabular}
	\caption{Trajectories of DAEs trained on the common data distribution \refeq{gmm4} (a GMM with non-uniform weight and uniform covariance).
	The \emb{gray lines} start from the regular grid. The \emb{colored lines} start from the samples drawn from $\data_0$.  Every lines are drawn into attractors.}
	  \label{fig:flow.gmm4}
\end{figure}

\begin{figure}[t]
	\centering
	\begin{tabular}{cc}
		\begin{minipage}{0.5\hsize}
			\centering
			Conti. DAE $\cdae_t$ \\ \vspace{-.5cm}
			\includegraphics[width=\textwidth]{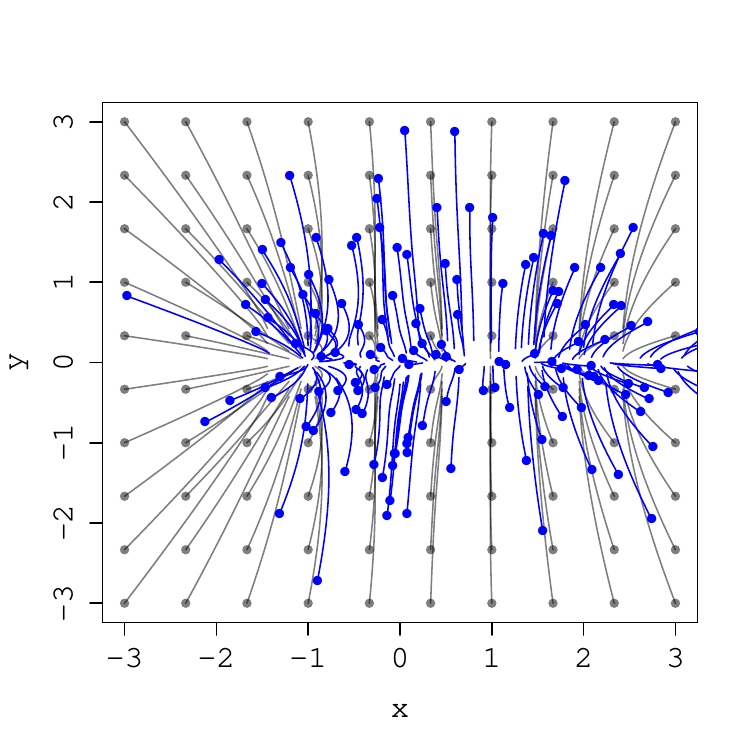}
		\end{minipage} &
		\begin{minipage}{0.5\hsize}
			\centering
			Shallow DAE $\gg_t$ \\ \vspace{-.5cm}
						\includegraphics[width=\textwidth]{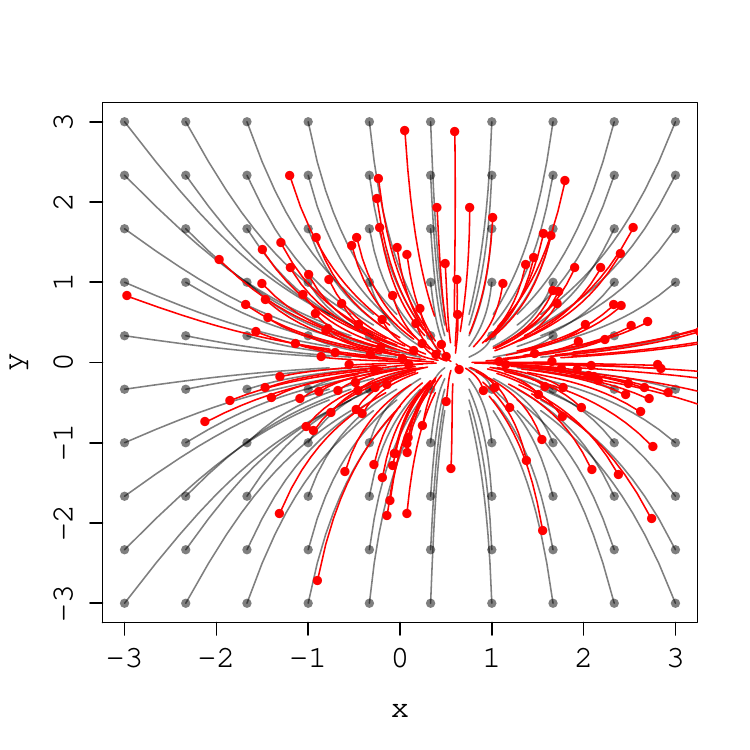}
		\end{minipage} \\
		\begin{minipage}{0.5\hsize}
			\centering
			Comp. DAEs $\gg_{0:L}^t \, (\dt=0.05)$ \\ \vspace{-.5cm}
						\includegraphics[width=\textwidth]{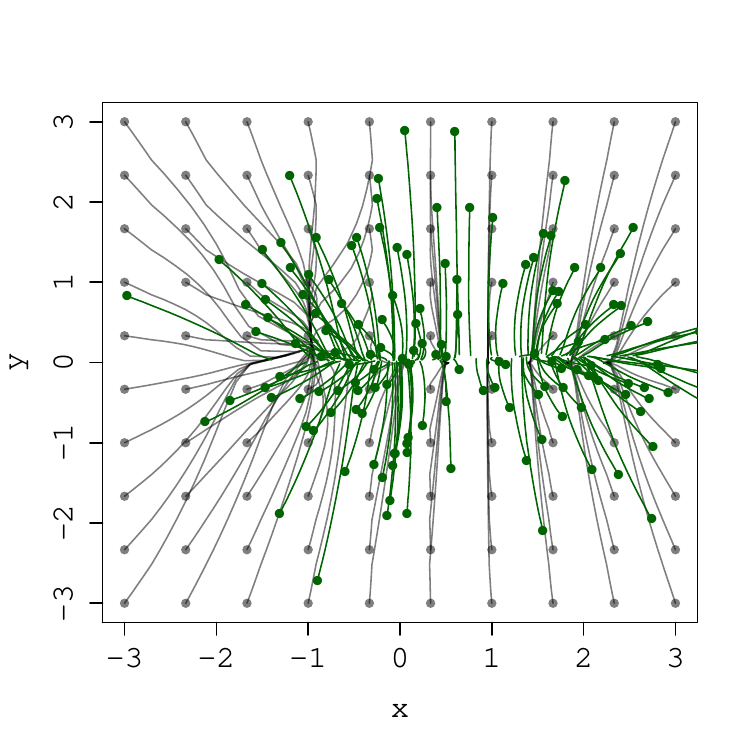}
		\end{minipage} &
		\begin{minipage}{0.5\hsize}
			\centering
		    Comp. DAEs $\gg_{0:L}^t \, (\dt=0.5)$ \\ \vspace{-.5cm}
						\includegraphics[width=\textwidth]{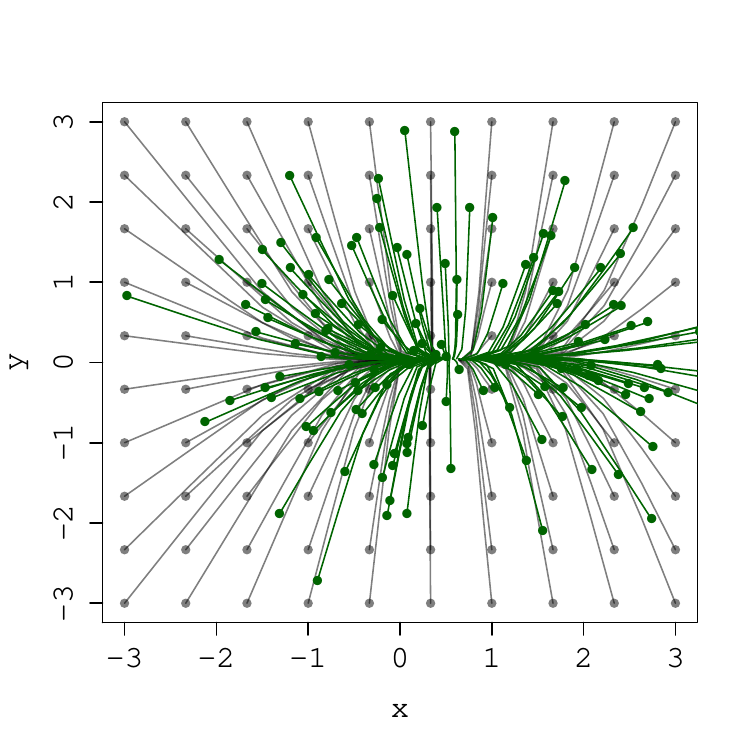}
		\end{minipage} \\
	\end{tabular}
	\caption{Trajectories of DAEs trained on the common data distribution \refeq{gmm5} (a GMM with non-uniform weight and covariance).
	The \emb{gray lines} start from the regular grid. The \emb{colored lines} start from the samples drawn from $\data_0$.  Every lines are drawn into attractors.}
		  \label{fig:flow.gmm5}
\end{figure}

\reffig{flow.gmm1}, \ref{fig:flow.gmm4}, and \ref{fig:flow.gmm5} compare the trajectories of four DAEs trained on the three common data distributions
\begin{align}
\data_0 &=
0.5 \,
N \left( [-1,0], \begin{bmatrix}
1 & 0 \\ 0 & 1
\end{bmatrix} \right)
+0.5 \,
N \left( [1,0], \begin{bmatrix}
1 & 0 \\ 0 & 1
\end{bmatrix} \right), \label{eq:gmm1} \\
\data_0 &=
0.2 \,
N \left( [-1,0], \begin{bmatrix}
1 & 0 \\ 0 & 1
\end{bmatrix} \right)
+0.8 \,
N \left( [1,0], \begin{bmatrix}
1 & 0 \\ 0 & 1
\end{bmatrix} \right), \label{eq:gmm4}\\
\data_0 &=
0.2 \,
N \left( [-1,0], \begin{bmatrix}
1 & 0 \\ 0 & 1
\end{bmatrix} \right)
+0.8 \,
N \left( [1,0], \begin{bmatrix}
2 & 0 \\ 0 & 1
\end{bmatrix} \right). \label{eq:gmm5}
\end{align}
respectively.

The transport maps for computing the trajectories are given by \refeq{eg.xst} for the shallow DAE and composition of DAEs.
For the continuous DAE, we compute the trajectories by numerically solving the definition of the continuous Gaussian DAE: $\dot{\xx} = \nabla \log \data_t(\xx)$.

In any case, the continuous DAE converges to an attractor at some $t >0$, but the shape of the attractors and the basins of attraction change according to the initial data distribution.
The shallow DAE converges to the origin as $t \to \infty$,
and the composition of DAEs plots a curve similar to that of the continuous DAE as $L$ tends to infinity, $\gg_{0:L}^t$.
In particular, in \reffig{flow.gmm4}, some trajectories of the continuous DAE intersect,
which implies that the velocity vector field $\vf_t$ is time-dependent.

\clearpage
\subsection{Numerical Example of Trajectories in Wasserstein Space}
We consider the space $\spQ$ of bivariate Gaussians:
\begin{align}
\spQ := \left\{
N \left( [0,0], \begin{bmatrix}
\sigma_1^2 & 0 \\ 0 & \sigma_2^2
\end{bmatrix} \right) \, \Bigg| \, \sigma_1, \sigma_2 > 0 \right\}.
\end{align}
Obviously, $\spQ$ is a $2$-dimensional subspace of $L^2$-Wasserstein space, and it is closed in the actions of the continuous DAE and shallow DAE because the pushforwards are given by \refeq{eg.mcp} and \refeq{eg.msp}, respectively.

We employ $(\sigma_1, \sigma_2)$ as the coordinate of $\spQ$. This is reasonable because, in this coordinate, the $L^2$-Wasserstein distance $W_2(\mu, \nu)$ between two points $\mu = (\sigma_1, \sigma_2)$ and $\nu = (\tau_1,\tau_2)$ is simply given by the ``Euclidean distance''
$W_2(\mu,\nu) = \sqrt{(\sigma_1 - \tau_1)^2 + (\sigma_2 - \tau_2)^2}$
\citep[see][for the proof]{Takatsu2011}.
The Shannon entropy is given by
\begin{align}
\ent(\sigma_1, \sigma_2) &= (1/2)\log | \diag[ \sigma^2_1, \sigma^2_2] | + const. = \log \sigma_1 + \log \sigma_2 + const.
\end{align}

\begin{figure}[h]
	\centering
	\begin{tabular}{cc}
		\begin{minipage}{0.5\hsize}
			\centering
			\includegraphics[width=\textwidth]{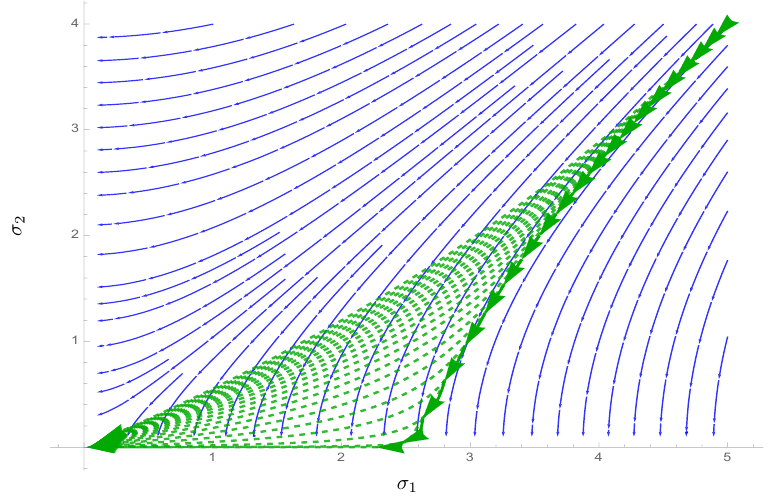}
		\end{minipage} &
		\begin{minipage}{0.5\hsize}
			\centering
			\includegraphics[width=\textwidth]{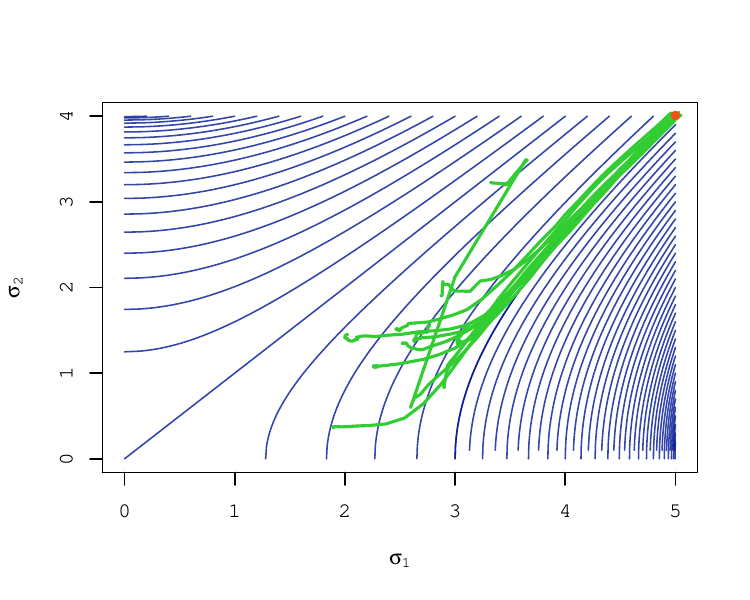}
		\end{minipage} \\
	\end{tabular}
    \caption{
		Trajectories of pushforward measures in a space $\spQ$ of bivariate Gaussians $N([0,0], \diag[ \sigma^2_1, \sigma^2_2 ])$.
		In \emb{both} sides, the \emb{blue} lines represent the Wasserstein gradient flow with respect to the Shannon entropy.
		The continuous Gaussian DAE $t \mapsto \cdae_{t \sharp} \data_0$ always coincides with the blue lines.
		In the \emb{left-hand side}, the \emb{dashed green} lines represent theoretical trajectories of the shallow DAE $t \mapsto \gg_{t \sharp} \data_0$
		and the \emb{solid green} line represents a theoretical trajectory of the composition of DAEs $t \mapsto \gg_{0:L \sharp}^t \data_0$.
		Both the green lines gradually leave the gradient flow.
		In the \emb{right-hand side}, the \emb{solid green} lines represent the trajectories of the composition of DAEs calculated by training real NNs ($10$ trials).
		In particular, in the early stage, the trajectories are parallel to the gradient flow.
		}
    \label{fig:aflow}
    \vspace{-1cm}
\end{figure}

\reffig{aflow} compares the trajectories of the pushforward by DAEs in $\spQ$.
In the left, we calculated the theoretical trajectories according to the analytic formulas \refeq{eg.mcp} and \refeq{eg.msp}.
In the right, we trained real NNs as the composition of DAEs according to the training procedure described in \refsec{compdae}.
Even though we always assumed the infinite number of hidden units and the infinite size of data set, 
the results suggest that our calculus is a good approximation to finite settings.

\section{Equivalence between Stacked DAE and Compositions of DAEs} \label{sec:stack.dae}
As an application of transport analysis, we shed light on the equivalence of the stacked DAE (SDAE) and the composition of DAEs (CDAE),
provided that the definition of DAEs is generalized to \emph{$L$-DAE}, which is defined below.
In SDAE, we apply the DAE to the features vectors obtained from the hidden layer of an NN to obtain higher-order feature vectors.
Therefore, the feature vectors obtained from the SDAE and CDAE are different from each other.
Nevertheless, we can prove that the trajectories generated by the SDAE and CDAE are \emph{topologically conjugate}, which means that there exists a homeomorphism between the trajectories.
Moreover, we can transform the trajectory of an SDAE into that of a CDAE by using a \emph{linear map}, which is obtained from the decoder of the SDAE.
Thus, we can synthesize the feature vectors of the SDAE by using CDAEs.

\subsection{Definitions}
To begin with, we introduce a generalized version of shallow DAE.
\begin{dfn}[$L$-DAE]
    Let $L$ be an elliptic operator on the domain $\Omega$ in $\RR^m$, $\data$ be a probability density on $\Omega$,
    and $D$ be a positive definite matrix. The $L$-DAE with diffusion coefficient $D$ and initial data $\data$ is defined by
    \begin{align}
        \id + t D \nabla \log e^{t L} \data, \quad t > 0.
    \end{align}
\end{dfn}
Here, $e^{tL}$ is the semigroup generated by the elliptic operator $L$. Specifically, let $\data_t := e^{tL} \data$; then, $\data_t$ satisfies the parabolic equation $\partial_t \data_t = L \data_t$.
The original Gaussian DAE corresponds to a special case when $D \equiv I$ and $L = \triangle$.

\begin{figure}[b]
    \begin{tabular}{cc}
        \begin{minipage}[t]{0.45\hsize}
        \centering
        \includegraphics[width=0.5\textwidth]{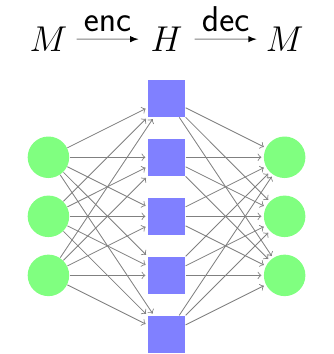}
        \end{minipage} &
        \begin{minipage}[t]{0.45\hsize}
        \centering
        \includegraphics[width=0.5\textwidth]{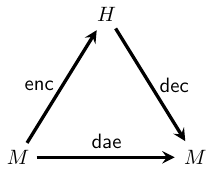}
        \end{minipage}\\
    \end{tabular}
	\caption{$\enc$ and $\dec$ correspond to the hidden layer and output layer, respectively.}
	\label{fig:daeshallow}
\end{figure}

By $\daenet$, we denote a DAE realized by a shallow NN (\reffig{daeshallow}). Specifically,
\begin{align}
    \daenet(\xx) = \sum_{j=1}^p \cc_j \sigma( \aa_j \cdot \xx - b_j ).
\end{align}
By $\enc$ and $\dec$, we denote the encoder and decoder of $\daenet$, respectively. Specifically,
\begin{align}
    \enc_j(\xx) &= \sigma( \aa_j \cdot \xx - b_j ), \quad j = 1, \ldots, p \\
    \dec(\zz) &= \sum_{j=1}^p \cc_j z_j,
\end{align}
where $z_j$ denotes the $j$-th element of $\zz = \enc(\xx)$.
Obviously, $\daenet = \dec \circ \enc$.

For the sake of simipicity, even though we introduced the finite number $p$ of hidden units, we assume that $p$ is large, and thus $\daenet$ approximately equals $L$-DAE for some $L$.

\subsection{Training Procedure of Stacked DAE (SDAE)}
Let $\ground := \RR^m$ be the space of input vectors with probability density $\data$, and let $\daenet : \ground \to \ground$ be a shallow NN with $p$ hidden units.
We assume that $\daenet$ is trained as the Gaussian DAE with $\data$, and it thus approximates the DAE $\id + t \nabla \log[ e^{t \triangle}\data ]$.
Let $\hidden := \RR^p$.
Then, the encoder and decoder of $\daenet$ are the maps $\enc : \ground \to \hidden$ and $\dec : \hidden \to \ground$, respectively.

\begin{figure}[h]
   \begin{tabular}{cc}
        \begin{minipage}[t]{0.45\hsize}
        \centering
        \includegraphics[width=0.5\textwidth]{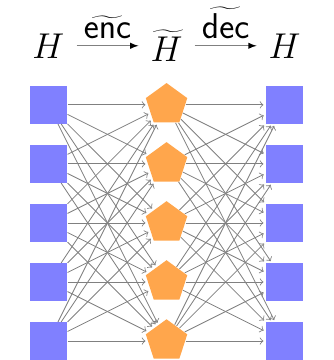}
        \end{minipage} &
        \begin{minipage}[t]{0.45\hsize}
        \centering
        \includegraphics[width=0.8\textwidth]{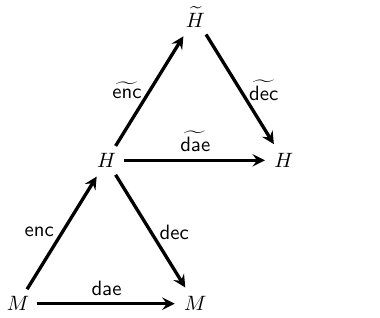}
        \end{minipage} \\
    \end{tabular}
	\caption{The (feature map of) SDAE $\tenc \circ \enc$ is built on the hidden layer.}
	\label{fig:daedeep1}
\end{figure}

In the SDAE, we apply the DAE to $\zz$.
Specifically, let $\tdata$ be the density of hidden feature vectors $\zz = \enc(\xx)$, and let $\tdaenet : \hidden \to \hidden$ be a shallow NN with $\tp$ hidden units,
\begin{align*}
 \tdaenet(\zz) := \sum_{\tj = 1}^\tp \tcc_\tj \sigma( \taa_\tj \cdot \zz - \tb_\tj ).
\end{align*}
We train $\tdaenet$ by using the Gaussian DAE with $\tdata$, where the network is decomposed as $\tdaenet = \tdec \circ \tenc$ with $\tenc : \hidden \to \thidden$ and $\tdec : \thidden \to \hidden$, and we obtain the feature vectors $\tzz := \tenc(\zz) \in \thidden = \RR^\tp$.
By iterating the stacking procedure, we can obtain more abstract feature vectors (\reffig{daedeep1}).

Technically speaking, $\tdata$ is (the density of) the pushforward $\daenet_{\sharp} \data$, and its support is contained in the image $\tground := \enc(\ground)$. In general, we assume that $\dim \tground (=\dim \ground) \leq \dim \hidden$; thus, the support of $\tdata$ is singular (i.e., the density vanishes outside $\tground$) (see \reffact{cov.singular} for further details).

\subsection{Topological Conjugacy}
\begin{figure}[t]
   \begin{tabular}{cc}
        \begin{minipage}[t]{0.45\hsize}
        \centering
        \includegraphics[width=\textwidth]{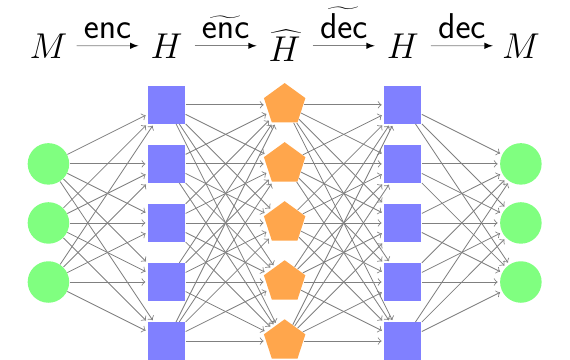}
        \end{minipage} &
        \begin{minipage}[t]{0.45\hsize}
        \centering
        \includegraphics[width=0.8\textwidth]{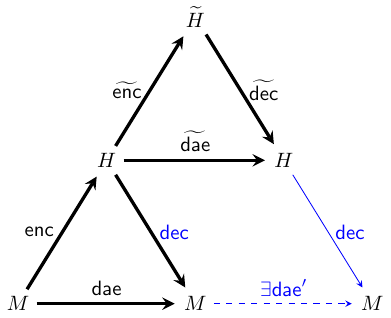}
        \end{minipage}\\
    \end{tabular}
	\caption{By reusing $\dec$, we can transform the SDAE $\tenc \circ \enc$ into a CDAE $\daenet' \circ \daenet$.}
	\label{fig:daedeep2}
\end{figure}

The transport map of the feature vector $\tenc \circ \enc : \ground \to \hidden \to \thidden$ is somewhat unclear.
According to \refthm{conjugate} and \ref{thm:stack.comp}, the transport map of $\tenc \circ \enc$ can be transformed or projected to the ground space $\ground$ by applying $\dec \circ \tdec$ (\reffig{daedeep2}). Specifically, there exists an $L$-DAE $\daenet' : \ground \to \ground$ such that
\begin{align}
    \dec \circ \tdec \circ \tenc \circ \enc = \daenet' \circ \daenet. \label{eq:eg.L2}
\end{align}

\begin{thm} \label{thm:conjugate}
    Let $\hidden$ and $\thidden$ be vector spaces, $\dim \hidden \geq \dim \thidden$,
    let $\ground_0$ be an $m$-dimensional smooth Riemannian manifold embedded in $\hidden$,
    and let $\data_0$ be a $C^2$ probability density on $\ground_0$.
    Let $\ff : \hidden \to \hidden$ be an $L_t$-DAE:
    \begin{align*}
        \ff := \id_\hidden + t D \nabla \log e^{t L_t} \data_0,
    \end{align*}
    with diffusion coefficient $D$ and time-dependent elliptic operator $L_t$ on $\hidden$,
    where $\nabla$ is the gradient operator in $\hidden$.

	Let $T : \hidden \to \thidden$ be a linear map.
    If $T|_\ground$ is injective, then
	there exists an $\tL_t$-DAE $\tff : \thidden \to \thidden$ with diffusion coefficient $\tD$ such that
	\begin{align}
		T \circ \ff|_{\ground} = \tff \circ T|_{\ground}. 	\label{eq:conjugate}
	\end{align}
\end{thm}

In other words, the following diagram commutes.
\begin{figure}[h]
	\centering
	\includegraphics[width=.5\textwidth]{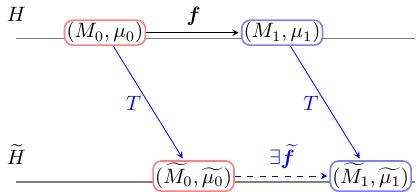}
\end{figure}
Here we denoted $\ground_1 := \ff( \ground_0 )$ and $\data_1 := \ff_\sharp \data_0$.
See \refapp{proof.conjugate} for the proof.
The statement is general in that the choice of a linear map $T$ is independent of the DAEs, as long as it is injective.

We note that 
the trajectory of the equivalent DAE $\tff$ may be complicated, because the ``equivalence'' we mean here is simply the topological conjugacy.
Actually, as the proof suggests, $\tD$ and $\tL_t$ contain the non-linearity of activation functions via the pseudo-inverse $T^\dag$ of $T$.
Nevertheless, $\tff$ may not be much complicated because it is simply a linear projection of the high-dimensional trajectory of $L_t$-DAE.
According to \refthm{dae.pfinit}, a Gaussian DAE solves backward heat equation (at least when $t \to 0$).
Hence, its projection to low dimension should also solve backward heat equation in low dimension spaces.

\subsection{Equivalence between SDAE and CDAE}
\begin{figure}[t]
	\centering
	\includegraphics[width=\textwidth]{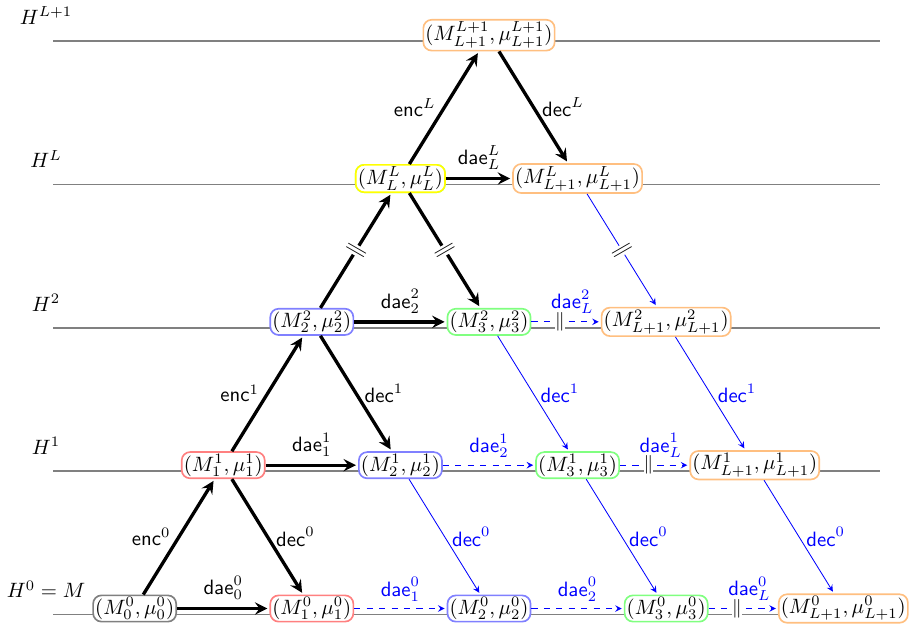}
	\caption{By using decoders, an SDAE is transformed or projected into a CDAE. The leftmost arrows correspond to the SDAE $\enc^{0:L}$,
the rightmost arrows correspond to the decoders $\dec^{L:0}$,
and the bottom arrows correspond to the CDAE $\daenet^0_L \circ \cdots \circ \daenet^0_0$.}
	\label{fig:pushforward}
\end{figure}

To clarify the statement, we prepare the notation.
\reffig{pushforward} summarizes the symbols and procedures.

First, we rewrite the input vector as $\zz^0$ instead of $\xx$, the input space as $\hidden^0 = \ground^0_0 (= \RR^m)$ instead of $\ground$, and the density as $\data^0_0$ instead of $\data$.
We iteratively train the $\ell$-th NN $\daenet^\ell_\ell : \hidden^\ell \to \hidden^\ell$ with a data distribution $\data^\ell_\ell$, obtain the encoder $\enc^\ell : \hidden^\ell \to \hidden^{\ell+1}$ and decoder $\dec^\ell : \hidden^{\ell+1} \to \hidden^\ell$, and update the feature $\zz^{\ell+1} := \enc^\ell(\zz^\ell)$, the image $\ground^{\ell+1}_{\ell+1} := \enc^\ell(\ground^\ell_\ell) \subset \hidden^{\ell+1}$, and the distribution $\data^{\ell+1}_{\ell+1} := (\enc^\ell)_\sharp \data^\ell_\data$.

For simplicity, we abbreviate
\begin{align*}
\enc^{\ell:n} &:= \enc^{n} \circ \cdots \circ \enc^\ell,\\
\dec^{n:\ell} &:= \dec^{\ell} \circ \cdots \circ \dec^n.
\end{align*}
In addition, we introduce auxiliary objects.
\begin{align*}
\ground_{\ell+1}^{n} &:= \dec^{\ell:n}( \ground_{\ell+1}^{\ell+1} ), \quad n = 0, \cdots, \ell \\
\data_{\ell+1}^{n} &:= \dec^{\ell:n}_\sharp \data_{\ell+1}^{\ell+1}, \quad n = 0, \cdots, \ell.
\end{align*}
By construction, $\ground_n^\ell$ is an at most $m$-dimensional submanifold in $\hidden^\ell$,
and the support of $\data_n^\ell$ is in $\ground_n^\ell$.

Finally, we denote the map $\daenet_n^\ell : \ground_n^\ell \to \ground_{n+1}^{\ell}$ that is (not ``trained by DAE'' but) defined by
\begin{align*}
\daenet_n^\ell := (\dec^{n:\ell} \circ \enc^{0:n}) \circ (\dec^{(n-1):\ell} \circ \enc^{0:(n-1)})^{-1} : \ground_{n}^\ell \to \ground_{n+1}^\ell.
\end{align*}
By \refthm{conjugate}, if $\daenet^{\ell+1}_n$ is an $L^{\ell+1}_n$-DAE, then $\daenet^{\ell}_n$ exists and it is an $L^{\ell}_n$-DAE.

\begin{thm} \label{thm:stack.comp}
	If every $\enc^\ell |_{\ground^{\ell}_\ell}$ is a continuous injection and every $\dec^\ell |_{\ground^{\ell+1}_n}$ is an injection,
	then
	\begin{align}
	\dec^{L:0} \circ \enc^{0:L} = \daenet^0_L \circ \cdots \circ \daenet^0_0.
	\end{align}
\end{thm}

\proofhere
By repeatedly applying the topological conjugacy in \refthm{conjugate},
\begin{align*}
\dec^\ell \circ \daenet^{\ell+1}_n = \daenet^\ell_n \circ \dec^\ell,
\end{align*}
we have
\begin{align*}
&\dec^{L:0} \circ \enc^{0:L} \\
&\ = \dec^{(L-2):0} \circ \dec^{L-1} \circ \daenet^L_L \circ \enc^{L-1} \circ \enc^{0:(L-2)} \\
&\ = \dec^{(L-2):0} \circ \daenet^{L-1}_L \circ \dec^{L-1} \circ \enc^{L-1} \circ \enc^{0:(L-2)} \\
&\ = \dec^{(L-2):0} \circ \daenet^{L-1}_L \circ \daenet^{L-1}_{L-1} \circ \enc^{0:(L-2)} \\
& \cdots \\
&\ = \daenet^0_{L} \circ \daenet^0_{L-1} \circ \cdots \circ \daenet^0_0. \qedhere
\end{align*}

\subsection{Numerical Example}
\begin{figure}[h]
	\centering
	\begin{tabular}{cc}
		\begin{minipage}{.5\hsize}
			\centering
			\includegraphics[width=\textwidth]{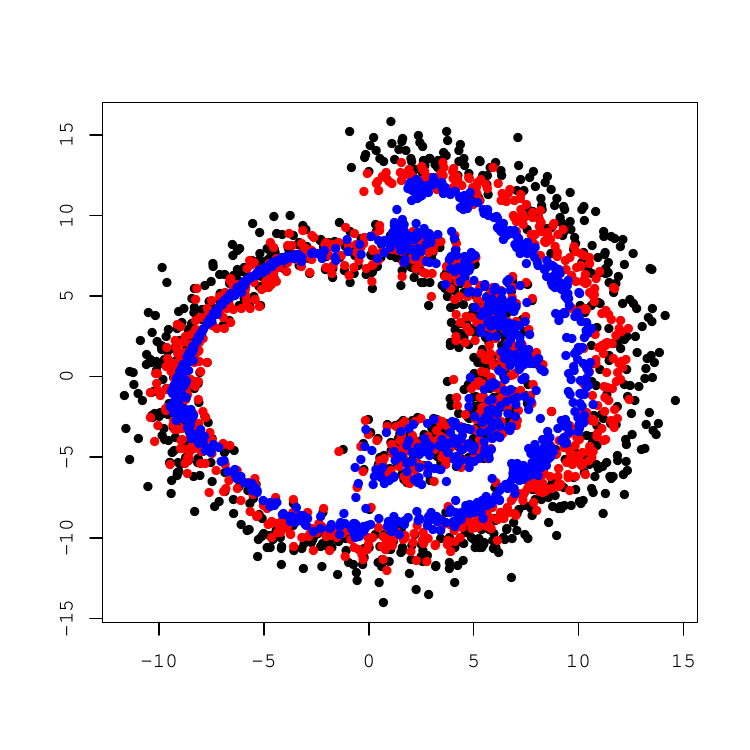}
		\end{minipage} &
		\begin{minipage}{.5\hsize}
			\centering
			\includegraphics[width=\textwidth]{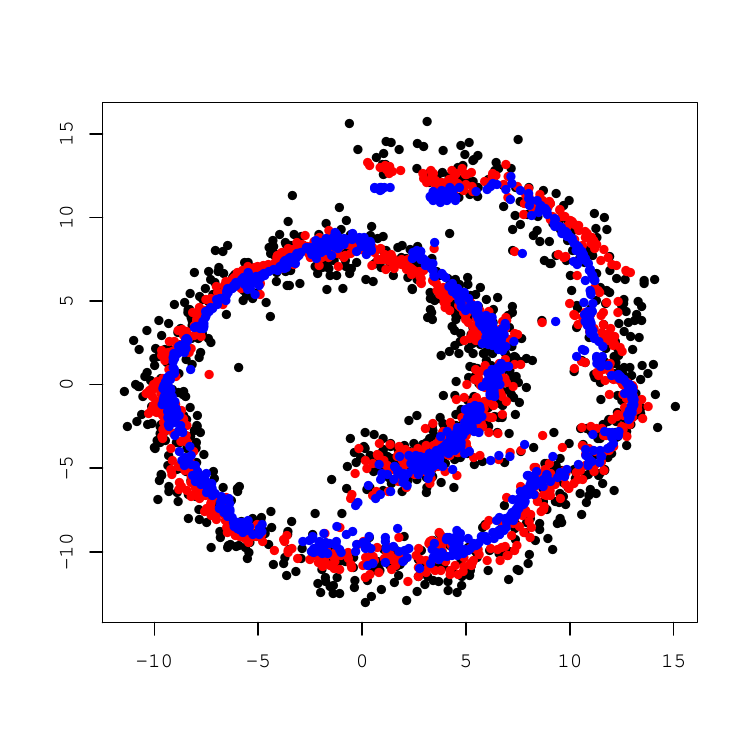}
		\end{minipage} \\
	\end{tabular}
	\caption{Typical transportation results of the $2$-dimensional swissroll data by an SDAE (left) and a CDAE (right).
	In \emb{both} the sides, the \emb{black} points represent the input vectors $\xx_0 \in \RR^2$, and the \emb{red} and \emb{blue} points represent the first and second transportation results $\xx_1$ and $\xx_2$, respectively.}
	\label{fig:swissroll.sdae.comdae}
\end{figure}

\reffig{swissroll.sdae.comdae} compares the transportation results of the $2$-dimensional swissroll data by the DAEs.
In both the cases, the swissroll becomes thinner by the action of transportation.
We remark that to test the topological conjugacy by numerical experiments is difficult.
Here, we display \reffig{swissroll.sdae.comdae} to see typical trajectories by an SDAE and a CDAE.

In the left-hand side, we trained an SDAE $\enc^1 \circ \enc^0$ by using real NNs. Specifically, we first trained a shallow DAE $\daenet^0_0$ on the swissroll data $\xx_0$.
Second, writing $\daenet^0_0 = \dec_0 \circ \enc_0$ and letting $\zz^1 := \enc_0(\xx_0)$, we trained a shallow DAE $\daenet^1_1$ on the feature vectors $\zz^1$.
Then, writing $\daenet^1_1 = \dec_1 \circ \enc_1$, we obtained $\xx_1 := \daenet^0_0(\xx_0)$ and $\xx_2 := \dec^0 \circ \dec^1 \circ \enc^1 \circ \enc^0$.
The black points represent the input vectors $\xx_0$, and the red and blue points represent the first and second transportation results $\xx_1$ and $\xx_2$, respectively. In other words, the distribution of $\xx_0, \xx_1$ and $\xx_2$ correspond to $\data_0^0, \data_1^0$ and $\data_2^0$ in \reffig{pushforward}, respectively.

In the right-hand side, we trained a CDAE $\daenet^1_0 \circ \daenet^0_0$ by using real NNs.
Specifically, we first trained a shallow DAE $\daenet^0_0$ on the swissroll data $\xx_0$.
Second, writing $\xx_1 := \daenet^0_0(\xx_0)$, we trained a shallow DAE $\daenet^0_1$ on the transported vectors $\xx^0_1$.
Then, we obtained $\xx_2 := \daenet^1_0(\xx_1) = \daenet^1_0 \circ \daenet^0_0 (\xx_0)$.
The black points represent the input vectors $\xx_0$, and the red and blue points represent the first and second transportation results $\xx_1$ and $\xx_2$, respectively.

\section{Integral Representation of the Flow Representation} \label{sec:intrep.flowrep}
In this section, we aim to develop the double continuum limit: a combination of the depth continuum limit, or the flow representation, and the width continuum limit, or the integral representation.

To facilitate visualization, we write the hidden parameters as $\ttheta$ instead of $(\aa,b)$,
the $k$-th element of the coefficient function as $\gamma(\ttheta, k)$ or $\gamma_k(\ttheta)$ instead of the boldface $\ggamma(\ttheta)$,
and the integral representation as
\begin{align}
S[\gamma_k](\xx) = \int \gamma(\ttheta, k) \sigma(\xx; \ttheta) \dd \ttheta. \label{eq:infinite.theta}
\end{align}
Furthermore, by using a singular measure $\gamma_k^p(\ttheta) := \sum_{j=1}^p c_{jk} \delta_{\ttheta_j}(\ttheta)$, we write an ordinary shallow NN as
\begin{align}
    S[\gamma_k^p](\xx) = \int \gamma^p(\ttheta, k) \sigma(\xx; \ttheta) \dd \ttheta = \sum_{j=1}^p c_{jk} \sigma( \xx; \ttheta_j ). \label{eq:finite.theta}
\end{align}
If there is no risk of confusion, we omit writing the superscript $p$. Specifically, we write ``$S[\gamma_k]$'' without distinction between an infinite NN \refeq{infinite.theta} and a finite NN \refeq{finite.theta}.

\subsection{Encoder and Decoder in the Integral Representation}
First, we consider a finite case.
Suppose that a shallow DAE is realized by a finite NN $\sum_{j=1}^p c_{jk} \sigma( \xx ; \ttheta_j )$. Then, the encoder is given by
\begin{align*}
    z(\ttheta_j) = \enc(\xx, \ttheta_j ) = \sigma( \xx ; \ttheta_j ), \quad j = 1, \ldots, p;
\end{align*}
and the decoder is given by
\begin{align*}
    \dec( \zz, k ) = \sum_{j=1}^p c_{jk} z( \ttheta_j ).
\end{align*}

Therefore, supposing that a shallow DAE is realized by $S[\gamma]$, the encoder and decoder in the integral representation are given by
\begin{align}
    \enc( \xx, \ttheta ) &:= \sigma( \xx ; \ttheta ), \\
    \dec( \zz, k ) &:= \int \gamma(\ttheta, k) z( \ttheta ) \dd \ttheta,
\end{align}
where ``the $\ttheta$-th element'' of $\zz$ is given by $z(\ttheta)$.

Next, we consider the stacked DAE built on $\zz$.
Suppose that the stacked DAE is realized by $S[\tgamma_\ttheta](\zz) = \int \tgamma(\oomega, \ttheta) \sigma( \zz ; \oomega ) \dd \oomega$; then, the encoder and decoder are given by
\begin{align}
    \tenc( \zz, \oomega ) &:= \sigma( \zz ; \oomega ), \\
    \tdec( \uu, \ttheta ) &:= \int \tgamma(\oomega, \ttheta) u( \oomega ) \dd \oomega,
\end{align}
where the $\oomega$-th element of $\uu$ is given by $u(\oomega)$, and the $\ttheta$-th element of $\oomega$ is given by $\omega(\ttheta)$.

In this notation, for example, the topological conjugacy \refeq{conjugate} claims that there exists $\gamma'$ such that
\begin{align}
    \int \gamma( \ttheta, k ) \int \tgamma( \oomega, \ttheta ) \sigma( \sigma( \xx ; \cdot ); \oomega ) \dd \oomega \dd \ttheta
    = \int \gamma'(\ttheta' , k) \sigma \left( \int \gamma(\ttheta, \cdot) \sigma( \xx ; \ttheta ) \dd \ttheta ; \ttheta' \right) \dd \ttheta'.
\end{align}

\subsection{Ridgelet Transform of Flows}
Let $\cmap_t : \RR^m \to \RR^m$ be a flow that satisfies $\cmap_t \circ \cmap_s = \cmap_{t+s}$. Then, the following formula holds:
\begin{align}
    \int R[\cmap_t]( \ttheta, k ) \sigma \left( \int R[\cmap_s]( \ttheta, \cdot ) \sigma( \xx; \ttheta' ) \dd \ttheta' \right) \dd \ttheta
    = \int R[\cmap_{t+s}]( \ttheta, k ) \sigma( \xx; \ttheta ) \dd \ttheta.
\end{align}
In other words, $S[R[\cmap_t]] \circ S[R[\cmap_s]] = S[R[\cmap_{t+s}]]$.
According to Barron's bound \citep[Cor.5.4]{Kurkova2012},
the discretization error $\| S[\gamma] - S[\gamma^p]\|_2$ between $S[\gamma]$ and $S[\gamma^p]$ is bounded by $\| \gamma \|_1 / \sqrt{p}$.
Hence, $\| R[\cmap_t] \|_1 + \| R[\cmap_s] \|_1 \leq \| R[\cmap_{t+s}] \|_1$ for some $t$ and $s$, which implies the expressive efficiency of the DNN.

Consider a special case when $\cmap : \RR^m \to \RR^m$ is given by the gradient of a potential function $V$. Specifically, $\cmap = \nabla V$.
We note that according to the polar decomposition theorem by \citet{Brenier1991}, any optimal transport map $\cmap_t : [0,1] \times \RR^m \to \RR^m$ can be written as $\cmap_t = \id + t \nabla U$ with some potential function $U$. Hence, by letting $V = | \cdot |^2/2 + U$, we can understand $\cmap := \cmap_1 = \nabla V$ as an optimal transport map.

Then, we have an integration-by-parts formula for the vector ridgelet transform.
\begin{thm}
Let $K \subset \RR^m$ be a compact set with smooth boundary $\partial K$. Given that a smooth scalar potential $V$ is supported in $K$,
the ridgelet transform of the potential vector field $\nabla V$ is calculated by
\begin{align}
R_\rho [\nabla V] (\aa,b)
&= - \aa R_{\rho'} [V] (\aa,b). \label{eq:vridge}
\end{align}
Here, $R_{\rho}$ and $R_{\rho'}$ denote the ridgelet transform with respect to $\rho$ and $\rho'$, respectively.
\end{thm}
\proofhere
\begin{align*}
R_\rho [\nabla V] (\aa,b)
&= \int_{K} \nabla V(\xx) \overline{\rho(\aa \cdot \xx - b)} \dd \xx \\
&=
\left[ \int_{\partial K} V(\xx) \overline{\rho(\aa \cdot \xx - b)} \nn(\xx) \dd S
- \aa \int_{K} V(\xx) \overline{\rho'(\aa \cdot \xx - b)} \dd \xx \right] \\
&= 0 - \aa \, R_{\rho'} [V] (\aa,b). \qedhere
\end{align*}
The left-hand side (LHS) of \refeq{vridge} denotes a vector ridgelet transform defined by element-wise mapping,
whereas the right-hand side (RHS) consists of a scalar ridgelet transform. We can understand the RHS given that the network shares common knowledge among element-wise tasks.

\subsection{Example: Autoencoder}
As the most fundamental transport map, we consider a smooth ``truncated'' autoencoder $\id_{r,\delta}$.
We denote by $\BB^m(\zz;r)$ a closed ball in $\RR^m$ with center $\zz$ and radius $r$.
We assume that $\id_{r,\delta}$ is (1) smooth, (2) equal to the identity map $\id$ when it is restricted to $\BB^m(r)$, and (3) truncated to be supported in $\BB^m(r+\delta)$ with a small positive number $\delta >0$.
Let $\potential_{r,\delta}$ be a smooth function that satisfies
\begin{align*}
V_{r,\delta}(\xx) := \begin{cases}
\frac{1}{2}|\xx|^2 & \xx \in \BB^m(0;r), \\
\mbox{(smooth map)} & \xx \in \BB(0; r+\delta) \setminus \BB(0;r), \\
0 & \xx \notin \BB^m(0;r+\delta),
\end{cases}
\end{align*}
and let
\begin{align*}
\id_{r,\delta} := \nabla V_{r,\delta}.
\end{align*}
Note that we can construct $\id_{r,\delta}$ and $\potential_{r,\delta}$ by using mollifiers; thus, such maps exist.

The ridgelet transform of the truncated autoencoder is given by
\begin{align}
R_\rho [\id_{r,\delta}] (\aa,b)
&\approx - K \aa \overline{ \rho'(-b) } \quad \mbox{as} \quad \delta \to 0 \label{eq:ridgeae}
\end{align}
with a certain constant $K$
(see \refapp{ridgeae} for the proof).
\clearpage
\section{Discussion}
We performed transport analysis of denoising autoencoders by introducing the flow representation.
The flow representation $\cmap_t$ is the depth continuum limit of a DNN, specified by an ODE with vector field $\vf_t$.
We interpreted an ordinary DNN $\gg_t$ as a transport map or an Euler broken line approximation of $\cmap_t$.
The advantages of the flow representation are that it provides the coordinate-free treatment of DNNs, avoiding the redundancy of the ordinary parametrization of DNNs, and that it facilitates our understanding of what DNNs do---it is the mass transportation controlled by $\vf_t$.
In addition, the advantage of the interpretation as mass transportation is that it can handle function composition.
In the transport analysis, we analyzed a flow in three aspects: a dynamical system described by a transport map or vector field, a pushforward measure described by a continuity equation, and Wasserstein gradient flow.
From the results in Wasserstein geometry, these aspects are closely connected, and the hyperparameter $\vf_t$ plays a central role as an intermediary.
For example, in the transport analysis of continuous DAEs, the potential functional of the Wasserstein gradient flow often facilitates our understanding of the flow because it is the Shannon entropy, which is a fundamental quantity in statistics and machine learning.

In \refsec{sdae} and \ref{sec:ddae}, we specified the transport maps of shallow, deep, and infinitely deep DAEs, and we gave their statistical interpretations. The shallow DAE is an estimator of the mean, while the deep DAE transports data points to decrease the Shannon entropy of the data distribution, which gives a partial answer to our research question ``what do hidden layers do?''
In \refsec{examples}, according to analytic and numerical experiments, we showed that deep DAEs converge faster and that the extracted features are different from each other, which gives a partial answer to the other question ``why do DNNs perform better?''
In \refsec{stack.dae}, we proved the equivalence between the stacked DAE and the composition of DAEs. Because of the peculiar construction, it is difficult to formulate and understand stacking. Nevertheless, by tracking the flow, we succeeded in formulating the stacked DAE.
In \refsec{intrep.flowrep}, we developed the double continuum limits, or the width continuum limit of the depth continuum limit. We presented some examples of the integral representation of the flow, such as encoder, decoder, and traditional autoencoder.

As a consequence of the equivalence, we can understand the so-called \emph{pre-training} and \emph{fine-tuning} strategy \citep{Bengio2006,Erhan2010} as an \emph{optimal control} problem.
Namely, write a DNN as a composite $\lmap \circ \cmap_t$ of classifier $\lmap : \RR^m \to [0,1]^n$ and flow $\cmap_t : \RR^m \to \RR^m$.
If $\cmap_t$ stays closer to the identity, $\lmap$ has to be more complex---and vice versa.
The pre-training regularizes the behavior of hidden layers by
\begin{align}
    \frac{\dd}{\dd t} \cmap_t( \xx ) = \vf_t( \cmap_t( \xx ) ), \quad \xx \in \RR^m, \, t > 0 \label{eq:pretrain}
\end{align}
and the fine-tuning specifies the relation between input and output by
\begin{align}
    \mbox{Minimize} \quad \EE_{X,Y}| Y - \lmap \circ \cmap_{t=1}(X) |^2 \quad \mbox{w.r.t NN } \lmap \circ \cmap_{t=1}. \label{eq:finetune}
\end{align}
Overall, we can understand the strategy as the control problem of system \refeq{pretrain} under restriction \refeq{finetune}.
Owing to ridgelet transform, shallow NNs are interpretable and principled.
Development of a ``solution operator'' to the control problem in the flow representation would open the way to the interpretable and principled alternative to DNNs.

\subsubsection*{Acknowledgement}
The authors thank the editor and reviewers for their supportive and insightful comments, which have improved the clarity of the argument significantly.
	The authors also acknowledge fruitful discussions with Dr.~Shotaro Akaho, Dr.~Kohei Yatabe, Dr.~Keisuke Yano, and Mr.~Kentaro Minami.
	This work was supported by JSPS KAKENHI (15J07517 and 18K18113).

\appendix

\section{Proof of \refthm{sonoda}} \label{app:proof.sonoda}

    By $\sploc(\RR^m)$ and $\spcomp(\RR^m)$, we denote the spaces of locally integrable functions and compactly supported smooth functions, respectively.
    We assume that $\gg : \RR^m \to \RR^m$ is locally integrable ($\sploc$).

\proofhere
	The proof follows from the calculus of variations.
	Let
	\begin{align*}
	L[\gg]
	&= \int_{\RR^m} \EE_{\eps}| \gg(\xx +\eps) - \xx |^2 \data_0(\xx) \dd \xx\\
	&= \int_{\RR^m} \EE_{\eps}[ | \gg(\xx') - \xx'+\eps |^2 \data_0(\xx'-\eps)] \dd \xx', \quad \xx' \gets \xx+\eps.
	\end{align*}
	Here, $L[\gg]$ always exists because $\gg \in \sploc(\RR^m) \subset L^2(\data * \noise)$.
	Then, for an arbitrary function $\func \in \spcomp(\RR^m)$, the first variation $\delta L [\func]$ is given by
	\begin{align*}
	\delta L[\func] &=\frac{d}{d s} L[\gg + s \func] \Big|_{s=0}\\
	&=  \int_{\RR^m} \frac{\partial}{\partial s} \EE_{\eps}[| \gg(\xx) + s \func(\xx) - \xx+\eps |^2 \data_0(\xx-\eps)] \dd \xx \Big|_{s=0} \\
	&=  2 \int_{\RR^m} \EE_{\eps}[( \gg(\xx) - \xx+\eps ) \data_0(\xx-\eps)] \func(\xx)\dd \xx.
	\end{align*}
	At a critical point $\gg^*$ of $L$, $\delta L[\func] \equiv 0$ for every $\func$. Hence,
	\begin{align*}
	\EE_{\eps}[( \gg^*(\xx) - \xx+\eps ) \data_0(\xx-\eps) ] =0, \quad \almost \xx,
	\end{align*}
	by the fundamental lemma of calculus of variations for integrable functions,
	and we have
	\begin{align*}
	\gg^*(\xx)
	&= \frac{\EE_{\eps}[(\xx-\eps ) \data_0(\xx-\eps)]}{\EE_\eps[\data_0(\xx-\eps)]} = \refeq{alain}\\
	&= \xx - \frac{\EE_{\eps}[\eps  \data_0(\xx-\eps)]}{\EE_\eps[\data_0(\xx-\eps)]} = \refeq{sonoda}.
	\end{align*}
	Note that $\gg^*$ attains the global minimum, because, for every function $\func$,
	\begin{align}
	L[\gg^* + \func]
	&= \int_{\RR^m} \EE_{\eps}[ | \eps - \EE_t[\eps|\xx] +\func(\xx) |^2 \data_0(\xx-\eps)] \dd \xx \nonumber \\
	&=
	\int_{\RR^m} \EE_{\eps}[ | \eps - \EE_t[\eps|x] |^2 \data_0(\xx-\eps)] \dd \xx
	+ \int_{\RR^m} \EE_{\eps}[ | \func(\xx) |^2 \data_0(\xx-\eps)] \dd \xx \nonumber \\
	&\qquad + 2 \int_{\RR^m} \EE_{\eps}[ ( \eps - \EE_t[\eps|\xx] )  \data_0(\xx-\eps)] \func(\xx) \dd \xx \nonumber \\
	&= L[\gg^*] + L[\func] + 2 \cdot 0 \geq L[\gg^*]. \qedhere
	\end{align}

\section{Proof of \reffact{contieq}} \label{app:proof.contieq}
For simplicity, we assume that $\gg, \vf$, and $\data$ are smooth.
See \citet[\S~8.1]{Ambrosio2008} for more generalized conditions on the continuity equation.

\proofhere
To facilitate visualization, we write $\gg(\xx,t), \vf(\xx,t) $, and $ \data(\xx,t)$ instead of $\gg_t(\xx), \vf_t(\xx)$, and $\data_t(\xx)$, respectively.

By definition,
\begin{align*}
\begin{cases}
\partial_t \gg(\gg(\xx,t),t) = \vf(\gg(\xx,t),t), & \xx \in \RR^m ,\ t>0 \\
\gg(\xx, 0) = 0, & \xx \in \RR^m.
\end{cases}
\end{align*}
In particular,
\begin{align*}
    \nabla \gg(\xx,0) = I.
\end{align*}
According to the change-of-variables formula, for any $\xx \in \RR^m$ and $t > s > 0$,
\begin{align*}
\data(\gg(\xx,t),t) \cdot |\nabla \gg(\xx,t)| = \data(\xx,s),
\end{align*}
where $| \wdot |$ denotes the determinant.
	
Take the logarithm on both sides and then differentiate with respect to $t$. Then, the RHS vanishes and the LHS is calculated as follows:
\begin{align*}
\partial_t \log[ \data(\gg(\xx,t),t) \cdot |\nabla \gg(\xx,t)|]
&= \frac{\partial_t [\data(\gg(\xx,t),t)]}{\data(\gg(\xx,t),t)} + \partial_t \log |\nabla \gg(\xx,t)|\\
&= \frac{(\nabla \data)(\gg(\xx,t),t) \cdot \partial_t \gg(\xx,t) + (\partial_t \data)(\gg(\xx,t),t)}{\data(\gg(\xx,t),t)} \\ & \qquad + \tr [ (\nabla \gg(\xx,t))^{-1} \nabla \partial_t \gg(\xx,t)],
\end{align*}
where the second term follows a differentiation formula by \citet[Eq.~43]{MatCookbook}
\begin{align*}
	\partial \log |J| = \tr[J^{-1} \partial J].
\end{align*}
	
By letting $t \to s + 0$,
\begin{align*}
\frac{\nabla \data(\xx,t) \cdot \vf(\xx,t) + (\partial_t \data)(\xx,t)}{\data(\xx,t)} + \tr [ \nabla \vf(\xx,t) ] = 0,
\end{align*}
which gives
\begin{align}
\partial_t \data(\xx,t) = - \nabla \cdot [\data(\xx,t) \vf(\xx, t)]. \qedhere
\end{align}

\section{Proof of \refthm{conjugate}} \label{app:proof.conjugate}

We show that the diagram commutes. Observe that $\ff = \id + t D \nabla \log e^{t L_t} \data$ is the sum of the present position $\id$ and the gradient $\nabla V$ of potential $V = \log e^{t L_t} \data$. We calculate the pushforward $\tnabla \tV$ and show that it coincides with $\tL_t$-DAE.
\begin{figure}[h]
	\centering
	\includegraphics[width=.5\textwidth]{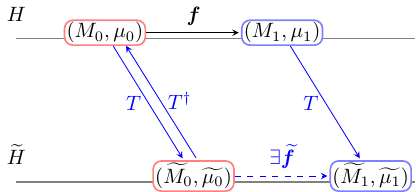}
\end{figure}

\proofhere
We suppose that $L_t$ is expressed as
\begin{align}
	L_t u &:= \aa_t^\top (\nabla^2 u) \aa_t + \bb_t^\top \nabla u + c_t u, \quad u \in C^2(\hidden)
\end{align}
and $T$ is expressed as
\begin{align}
T(\zz) = A \zz
\end{align}
with a matrix $A$.

By the assumption that the restriction $T|_{\ground_0}$ is injective, it has a left inverse $T^\dag$ such that $T^\dag \circ T|_{\ground_0} = \id_{\ground_0}$.
Note that it is not a linear map but an abstract nonlinear map, which means that there is no matrix $A$ that realizes $T^\dag$.

\subsection*{Step.~1}
We show that
\begin{align}
    T \circ \ff \circ T^\dag = \id + t \tD \tnabla \tV \quad \mbox{in } \tground_0
\end{align}
where $\tD = A D A^\top$ and $\tV = V \circ T^\dag$.

For an arbitrary $U \in C^2(\ground_0)$, write $T_* U := U \circ T^\dag \in C^2(\tground_0)$, and
\begin{align}
    \nabla U( T^\dag(\xx) ) = A^\top \tnabla T_* U( \xx ), \quad \xx \in \tground_0
\end{align}
because the $i$-th element of $\tnabla T_*U$ is given by
\begin{align*}
    \frac{\partial U \circ T^\dag}{\partial x_i}(\xx) = \sum_{p}\frac{\partial U}{\partial z_p}( T^\dag(\xx) ) \frac{\partial T^\dag_p}{\partial x_i}(\xx).
\end{align*}
Thus, the $q$-th element of $A^\top \tnabla T_*U$ is given by
\begin{align*}
    \sum_{i} A_{iq} \frac{\partial U \circ T^\dag}{\partial x_i}(\xx)
    &= \sum_{p}\frac{\partial U}{\partial z_p}( T^\dag(\xx) ) \sum_{i} A_{iq} \frac{\partial T^\dag_p}{\partial x_i}(\xx) \\
    &= \sum_{p}\frac{\partial U}{\partial z_p}( T^\dag(\xx) ) \delta_{pq} \\
    &= \frac{\partial U}{\partial z_q}( T^\dag(\xx) ).
\end{align*}
Therefore, by substituting $U$ with $V = \log e^{t L_t} \data_0$,
\begin{align*}
    T \circ \ff \circ T^\dag (\xx)
    &= \xx + A ( t D \nabla V ( T^\dag(\xx) ) )  \\
    &= \xx + t A  D A^\top \tnabla T_*V (\xx) ) \\
    &= \xx + t \tD \tnabla \tV (\xx).
\end{align*}

\subsection*{Step.~2}
We show that
\begin{align}
    \tV = \log e^{t \tL_t} \tdata_0 + (const.), \quad \mbox{in } \ground_0
\end{align}
where
\begin{align}
\tL_t \tu &:= \taa_t^\top (\tnabla^2 \tu) \taa_t + \tbb_t^\top \tnabla \tu + \tc_t \tu, \quad \tu \in C^2(\thidden)
\end{align}
with $\taa_t = A \aa_t \circ T^\dag, \tbb_t = A \bb_t  \circ T^\dag$, and $\tc_t = c_t  \circ T^\dag$.

Let
\begin{align}
    u_t := e^{t L_t} \data_0.
\end{align}
By the definition of semigroup $e^{t L_t}$, $u_0 = \data_0$ and $\partial_t u_t = L_t u_t$
(however, $u_1$ is different from $\data_1$).

Given $u_t$, let
\begin{align}
    \tu_t := T_\sharp u_t.
\end{align}
According to the change-of-variables formula \refeq{cov.singular},
\begin{align}
    \tu_t = [A]^{-1} T_* u_t,
\end{align}
where $[A] := \sqrt{ \det| A^\top A | }$ and $T_* u_t := u_t \circ T^\dag$.
In particular, $\tu_0 = \tdata_0$ and $\log \tu_t = \tV$.

Furthermore,
\begin{align}
    \partial_t \tu_t = \tL_t \tu_t, \quad \mbox{in } \tground_0,
\end{align}
because
\begin{align*}
    \partial_t \tu_t(\xx)
    &= [A]^{-1} \partial_t [ u_t( T^\dag(\xx) ) ] \\
    &= [A]^{-1} L_t [ u_t ](T^\dag(\xx)),
\end{align*}
and
\begin{align*}
&[A]^{-1} \aa_t(T^\dag(\xx))^\top (\nabla^2 u_t(T^\dag(\xx))) \aa_t(T^\dag(\xx)) \\
&\quad = \aa_t(T^\dag(\xx))^\top (A^\top \tnabla^2 [[A]^{-1} T_*u_t](\xx) A) \aa_t(T^\dag(\xx))\\
&\quad = \taa_t(\xx)^\top (\tnabla^2 [\tu_t](\xx)) \taa_t(\xx), \\
&[A]^{-1} \bb_t(T^\dag(\xx))^\top \nabla u_t(T^\dag(\xx)) \\
&\quad = \bb_t(T^\dag(\xx))^\top A^\top \tnabla [[A]^{-1} T_*u_t](\xx) \\
&\quad = \tbb_t(\xx)^\top \tnabla \tu_t(\xx), \\
&[A]^{-1} c_t(T^\dag(\xx)) u_t(T^\dag(\xx)) \\
&\quad = \tc_t(\xx) \tu_t(\xx).
\end{align*}
Thus,
\begin{align*}
    \partial_t \tu_t(\xx)
    = [A]^{-1} L_t [ u_t ](T^\dag(\xx)) = \tL_t \tu_t(\xx).
\end{align*}

Hence, $\tu_t$ is the solution of the initial value problem $\partial_t \tu_t = \tL_t \tu_t$ with $\tu_0 = \tdata_0$.
By the uniqueness of the solution, $\tu_t = e^{t \tL_t} \tdata_0$. On the other hand, $\log \tu_t = \tV$. Therefore, $\tV = \log \tu_t = e^{t \tL_t} \tdata_0$.

To sum up the two steps,
\begin{align*}
    T \circ \ff \circ T^\dag = \id + t \tD \tnabla \log e^{t \tL_t} \tdata_0 =: \tff,
\end{align*}
and we have the topological conjugacy
\begin{align}
    T \circ \ff = \tff \circ T. \qedhere
\end{align}

\section{Proofs for Analytic Examples} \label{app:anal.examples}

\subsection{Univariate Normal Distribution}
We calculate the case for a univariate normal distribution $N(m_0, \sigma_0^2)$.

\subsubsection{Shallow DAE} 

We show that
\begin{align}
g_t(x) &= \frac{\sigma_0^2}{\sigma_0^2 + t}x + \frac{t}{\sigma_0^2 + t} m_0, \tag{\ref{eq:eg.ust}}\\
\data_t &= N \left( m_0, \frac{\sigma_0^2}{(1 + t /\sigma_0^2)^2} \right). \tag{\ref{eq:eg.usp}} 
\end{align}

\begin{proof}
The proof is immediate from \refeq{gdae.sonoda}.
First, write $\heat_t(x,y) = (4 \pi t)^{-1/2} \exp( -|x-y|^2/4t )$,
\begin{align*}
\heat_{t/2} * N(m_0, \sigma_0^2) = N(m_0, \sigma_0^2 + t).
\end{align*}
Hence,
\begin{align*}
g_t(x)
= x + t \nabla \log[ N(m_0, \sigma_0^2+t)]
= \frac{\sigma_0^2}{\sigma_0^2 + t}x + \frac{t}{\sigma_0^2 + t} m_0.
\end{align*}
As $g_t$ is affine, the pushforward is immediate.
\end{proof}

\subsubsection{Continuous DAE} 

We show that
\begin{align}
g_t(x) &= \sqrt{ 1 - 2t/\sigma_0^2} (x-m_0) + m_0, \tag{\ref{eq:eg.uct}} \\
\data_t &= N( m_0,\sigma_0^2 - 2 t), \quad 0 \leq t < \sigma_0^2/2. \tag{\ref{eq:eg.ucp}} 
\end{align}

\begin{proof}[$\data_t$]
Write the pushforward as $N(m_t, \sigma_t^2)$.
By using the heat kernel $\heat_t(\xx,\yy) = (4 \pi t)^{-m/2} \exp( -|\xx-\yy|^2/4t )$, for some $\tbound > 0$,
\begin{align*}
N(m_t, \sigma_t^2)
&= \heat_{\tbound - t} * N(m_\tbound, \sigma_\tbound^2) \\
&= N(m_\tbound, \sigma_\tbound^2+2(\tbound - t)).
\end{align*}
By eliminating $\tbound$ by the initial conditions, we have
\begin{align*}
N(m_t, \sigma_t^2) &= N(m_0, \sigma_0^2 - 2t).
\end{align*}
By the positivity of $\sigma_t^2$, we can determine the largest possible $\tbound$ as $\tbound = \sigma_0^2/2$.
\end{proof}

\begin{proof}[$g_t$]
Fix an arbitrary point $x_0$.
Write $x_t := g_t(x_0)$ and $\dot{x_t} := \partial_t g_t(x_0)$.
Recall that $\dot{m_t} \equiv 0$, because $m_t$ is a constant.
According to \refeq{conti},
\begin{align*}
\dot{x_t}
&= -\frac{x_t - m_t}{\sigma_t^2}.
\end{align*}
By dividing both sides by $x_t$ and integrating them,
\begin{align*}
\log \Big|\frac{x_t - m_t}{x_0 - m_0}\Big|
&= -\int_0^t \frac{\dd s}{\sigma_s^2} \\
&= \frac{1}{2} \int_0^t \frac{\dd s}{s-\tbound} \\
&= \frac{1}{2} \log \Big| \frac{\tbound - t}{\tbound} \Big|,
\end{align*}
which concludes the proof.
\end{proof}

\subsection{Multivariate Normal Distribution}

We calculate the case for a multivariate normal distribution $N(\mm_0, \Sigma_0)$.
\subsubsection{Shallow DAE}

We show that
\begin{align}
\gg_t(\xx) &= (I + t \Sigma_0^{-1})^{-1}\xx + (I + t^{-1}\Sigma_0)^{-1} \mm_0, \tag{\ref{eq:eg.mst}}\\ 
\data_t &= N( \mm_0, \Sigma_0(I + t \Sigma_0^{-1})^{-2} ).\tag{\ref{eq:eg.msp}}
\end{align}

\begin{proof}
Calculate \refeq{gdae.sonoda} directly as in the univariate case.
First, by writing $\heat_t(\xx,\yy) = (4 \pi t)^{-m/2} \exp( -|\xx-\yy|^2/4t )$,
\begin{align*}
\heat_{t/2} * \, N(\mm_0, \Sigma_0)
&= N(\mm_0, \Sigma_0 + tI).
\end{align*}
Hence,
\begin{align*}
\gg_t(\xx)
&= \xx + t \nabla \log[ N( \mm_0, \Sigma_0 + tI ) ] \\
&= \xx + t \nabla \left[ -\frac{1}{2} (\xx - \mm_0)^\top ( \Sigma_0 + tI )^{-1} (\xx - \mm_0) \right] \\
&= (I + t \Sigma_0^{-1})^{-1} \xx + (I + t^{-1} \Sigma_0)^{-1} \mm_0.
\end{align*}
As $\gg_t$ is affine, the pushforward is immediate.
\end{proof}

\subsubsection{Continuous DAE}

We show that
\begin{align}
\gg_t(\xx) &= \sqrt{ I - 2 t \Sigma_0^{-1 }}(\xx - \mm_0) + \mm_0, \tag{\ref{eq:eg.mct}}\\
\data_t &= N( \mm_0, \Sigma_0 - 2 t I ). \tag{\ref{eq:eg.mcp}}
\end{align}

\begin{proof}
Write $\heat_t(\xx,\yy) = (4 \pi t)^{-m/2} \exp( -|\xx-\yy|^2/4t )$, and
recall that $\heat_{t} * N(\mm,\Sigma) = N(\mm,\Sigma + 2 t I)$.
Thus, the pushforward $N(\mm_t, \Sigma_t)$ is obtained as follows in a manner similar to the univariate case.
\begin{align*}
N(\mm_t,\Sigma_t) &= N \left( \mm_0, \Sigma_0 - 2 t I \right).
\end{align*}

Suppose that $\gg_t(\xx)$ is an affine transform $A_t (\xx-\mm_0)+\mm_0$ analogous to the univariate case.
Recall that, if $X \sim N(\mm,\Sigma)$, then $AX+b \sim N(A\mm+b, A \Sigma A^\top)$.
Hence, for our case, $\Sigma_t = A_t \Sigma_0 A_t^\top$ and we can determine
\begin{align*}
A_t = \sqrt{ \Sigma_t \Sigma_0^{-1} } = \sqrt{ I - 2 t \Sigma_0^{-1} }.
\end{align*}

Finally, we check whether  $\gg_t$ satisfies \refeq{conti}.
As $\Sigma_0$ is symmetric, we can always diagonalize $\Sigma_0 = U D_0 U^\top$ with an orthogonal matrix $U$ and a diagonal matrix $D_0$.
Observe that with the same $U$, we can simultaneously diagonalize $\Sigma_t$ and $A_t$ as
\begin{align*}
\Sigma_t &= U D_t U^\top, \quad D_t := D_0 - 2t I \\
A_t &= U D_t^{1/2} D_0^{-1/2} U^\top.
\end{align*}

Without loss of generality, we can assume that $U=I$; therefore, $\Sigma_t$ and $A_t$ are diagonal and $\mm_t \equiv 0$.
Fix an index $j$ and denote the $j$-th diagonal element of $\Sigma_t$ and $A_t$ by $\sigma_t^2$ and $a_t$, respectively.
Then, our goal is reduced to showing that $\partial_t [a_t x] = \nabla \log \data_t( a_t x )$ for every fixed $x \in \RR$.

By definition,
\begin{align*}
\sigma_t^2 &= \sigma_0^2 - 2 t, \\
a_t &= \sigma_t \sigma_0^{-1} = \sqrt{1- 2 t \sigma_0^{-2}}.
\end{align*}
Thus, the LHS is
\begin{align*}
\partial_t [a_t x] &= -\frac{1}{\sigma_0 \sqrt{\sigma_0^2-2t}} x = - \sigma_0^{-1}\sigma_t^{-1} x,
\end{align*}
and the RHS is
\begin{align*}
\nabla \log \data_t( a_t x )
 &= - \frac{a_t x}{\sigma_t^2} = -\sigma_0^{-1}\sigma_t^{-1} x.
\end{align*}
Hence, the LHS equals the RHS.
\end{proof}

\subsection{Mixture of Multivariate Normal Distributions}
We calculate the case for the mixture of multivariate normal distributions $\sum_{k=1}^K w_k N\left( \mm_k, \Sigma_k \right)$, with the assumption that it is \emph{well separated} (see \refsec{eg.x} for the definition).

\subsubsection{Shallow DAE}

We show that
\begin{align}
\gg_{t}(\xx) &= \sum_{k=1}^K \gamma_{kt}(\xx) \left\{ (I+ t \Sigma_k^{-1})^{-1}\xx + (I+t^{-1}\Sigma_k)^{-1}\mm_k \right\},\tag{\ref{eq:eg.xst}}\\ 
\data_t & \approx \sum_{k=1}^K w_k N( \mm_k, \Sigma_k(I + t \Sigma_k^{-1})^{-2} ), \quad \mbox{if well separated}\tag{\ref{eq:eg.xsp}}
\end{align}
with the responsibility function
\begin{align}
\gamma_{kt}(\xx) &:= \frac{w_k N(\xx;\mm_k , \Sigma_k + t I)}{\sum_{k=1}^K w_k N(\xx;\mm_k , \Sigma_k + t I)}. \tag{\ref{eq:eg.xsr}} 
\end{align}

\begin{proof}
Directly calculate \refeq{gdae.sonoda}. By the linearity of the heat kernel,
\begin{align*}
\gg_t
&:= \id + t \sum_{k=1}^K \frac{ w_k \nabla N( \mm_k, \Sigma_k + t I) }{\sum_{k=1}^K w_k N(\mm_k, \Sigma_k + tI)}, \\
&= \id + \sum_{k=1}^K \frac{ w_k N( \mm_k, \Sigma_k + t I) }{\sum_{k=1}^K w_k N( \mm_k, \Sigma_k + tI)} \cdot t \nabla \log N(\mm_k, \Sigma_k + t I), \\
&= \id + \sum_{k=1}^K \gamma_{kt} (\gg_{kt} - \id),\\
&= \sum_{k=1}^K \gamma_{kt} \gg_{kt},
\end{align*}
where $\gg_{kt}$ exactly coincides with the flow induced by the individual $k$-th component.

To calculate the pushforward, we introduce some auxiliary variables.
Write $w(k) := w_k, \ \gamma(k \mid \cdot) := \gamma_{kt}(\cdot)$ and
\begin{align*}
\data_t( \cdot \mid k) &:= N(\mm_k, \Sigma_k + t I), \\
\data_t &:= \sum_k w(k) \data_t(\cdot \mid k).
\end{align*}
Let $\tau_k( \cdot \mid \xx)$ be a probability measure that satisfies
\begin{align*}
\int_\ground \tau_k(y \mid \xx) \data_0(\xx \mid k) \dd \xx = \data_t(y \mid k).
\end{align*}
Note that $\tau_k$ is not unique.
Recall that by definition, if $X \sim \data_0(\cdot \mid k)$, then $Y=\gg_{kt}(X) \sim \data_t(\cdot \mid k)$.
Hence, $\tau_k$ is a stochastic alternative to $\gg_{kt}$.

Consider a probability measure
\begin{align*}
\sigma(\cdot \mid \xx) := \sum_{k=1}^K \gamma(k \mid \xx) \tau_k( \cdot \mid \xx ).
\end{align*}
Clearly, this is a stochastic alternative to $\gg_{t}$.
We show that
\begin{align*}
\int_\ground \sigma(y \mid \xx) \data_0(\xx) \dd \xx \approx \data_t(y).
\end{align*}
The LHS is reduced to
\begin{align}
\int_\ground \sigma(y \mid \xx) \data_0(\xx) \dd \xx
&= \int_\ground \sum_{k=1}^K \gamma(k \mid \xx) \tau_k(y \mid \xx) \sum_{\ell} w(\ell) \data_0(\xx \mid \ell) \dd \xx \nonumber \\
&= \sum_{\ell} w(\ell) \sum_{k=1}^K \int_\ground \gamma(k \mid \xx) \tau_k(y \mid \xx)  \data_0(\xx \mid \ell) \dd \xx. \label{eq:mgauss.disc.pf.lhs}
\end{align}
Suppose that $\gamma(k \mid \xx)$ is an indicator function of a domain $\Omega_k$, where $\int_{\Omega_k} \data_0(\cdot \mid k) \approx 1$.
Then,
\begin{align*}
\refeq{mgauss.disc.pf.lhs}
&\approx \sum_{\ell} w(\ell) \int_{\Omega_\ell} \tau_k(y \mid \xx)  \data_0(\xx \mid \ell) \dd \xx \\
&\approx \sum_{\ell} w(\ell) \data_t(y \mid \ell)
= \data_t(y).
\end{align*}
This concludes the claim.
\end{proof}

\subsubsection{Continuous DAE}

We show that
\begin{align}
\gg_t(\xx) &\approx \sqrt{ I - 2 t \Sigma_k^{-1} }\left( \xx - \mm_k \right) + \mm_k, \quad \xx \in \Omega_k, \mbox{ if well separated}\tag{\ref{eq:eg.xct}} \\
\data_t &= \sum_{k=1}^K w_k N \left( \mm_k, \Sigma_k - 2 t I \right),\tag{\ref{eq:eg.xcp}}
\end{align}
with the responsibility function
\begin{align}
\gamma_{kt}(\xx) &:= \frac{w_k N(\xx;\mm_k , \Sigma_k - 2 t I)}{\sum_{k=1}^K w_k N(\xx;\mm_k , \Sigma_k - 2 t I)}. \tag{\ref{eq:eg.xcr}} 
\end{align}

\proofhere
The pushforward is immediate by the linearity of the heat kernel.
The dynamical system \refeq{conti} for our case is reduced to
\begin{align*}
\partial_t \gg_t(\xx) &= -\sum_{k=1}^K \gamma_{kt} \circ \gg_t(\xx) ( \Sigma_k - 2 t I )^{-1}( \gg_t(\xx) - \mm_k ).
\end{align*}

By the assumption that $\data_0$ is well separated,
we can take an open neighborhood $\Omega_k$ of $\mm_k$ and an open time interval $I$ that contains $t$
 such that $\gamma_{kt} \circ \gg_t(\xx) \equiv 1$ for every $(\xx,t) \in \Omega_k \times I$.
In this restricted domain, the dynamical system is reduced to a single-component version:
\begin{align*}
\partial_t \gg_t(\xx) &= - ( \Sigma_k - 2 t I )^{-1}( \gg_t(\xx) - \mm_k ), \quad (\xx,t) \in \Omega_k \times I.
\end{align*}
According to the previous results, we have exactly
\begin{align*}
\gg_t(\xx) &= \sqrt{ I - 2 t \Sigma_k^{-1}} ( \xx - \mm_k ) + \mm_k, \quad (\xx,t) \in \Omega_k \times I. \qedhere
\end{align*}

\section{Proof of \refeq{ridgeae}} \label{app:ridgeae}

    Let $\delta \to 0$. Then, the ridgelet transform of the truncated autoencoder $\id_{r,\delta}$ is given by
    \begin{align}
    R_\rho[\id_{r,0}](\aa, b)
    &=  -\frac{A_{m-1}}{2(m+1)} \int_{|p| < r} (r^2-p^2)^\frac{m-1}{2} \left\{ \frac{2}{m-1} p^2 + r^2  \right\} \overline{\rho'(|\aa|p-b)} \aa \dd p\\
    & \approx -K \aa \overline{\rho'(-b)},
    \end{align}
    where $A_{m-1} := \frac{2 \pi^{\frac{m-1}{2}} }{ \Gamma \left( \frac{m-1}{2} \right)}$ is the surface area of $\Sph^{m-1}$,
    and $K$ is given by \refeq{def.K}.

\proofhere
    Let $\delta \to 0$. Then, the connecting annulus $\BB(0; r+\delta) \setminus \BB(0;r)$ vanishes as follows:
    \begin{align*}
    R_\rho [\id_{r,\delta}] (\aa, b)
    &= - \aa R_{\rho'} [V_{r,\delta}] (\aa, b) \\
    & \to -\aa \int_{\BB^m(r)} \frac{1}{2}|\xx|^2 \overline{\rho'(\aa \cdot \xx - b)} \dd \xx \\
    &= - \aa R_{\rho'} [V_{r,0}] (\aa, b).
    \end{align*}
    Hence, we omit considering the annulus.

    In the following, we use a spherical coordinate defined by
    \begin{align*}
    \uu := \aa/|\aa|, \quad \alpha := 1/|\aa|, \quad \beta := b/|\aa|,
    \end{align*}
    where $\uu \in \Sph^{m-1}$ denotes the direction, $\alpha \in \RR_+$ denotes the scale, and $\beta \in \RR$ denotes the (scaled) shift parameters.

    The ridgelet transform in the spherical coordinate \citep{Sonoda2015} is given by
    \begin{align*}
    R_\rho f(\uu/\alpha,\beta/\alpha) = \int_\RR \rad [f](\uu,p) \overline{\rho_\alpha(p - \beta)} \dd p,
    \end{align*}
    where $\rad [f](\uu,p)$ denotes the Radon transform
    \begin{align*}
    \rad [f](\uu,p) := \int_{(\RR \uu)^\perp} f(p\uu+\yy) \dd \yy
    \end{align*}
    of the function $f \in L^1(\RR^m)$ at direction $\uu \in \Sph^{m-1}$ and position $p \in \RR$,
    and
    \begin{align*}
    \rho_\alpha(p) := \rho(p/\alpha).
    \end{align*}

    The Radon transform $\rad[V_{r,0}](\uu,p)$ for $|p|<r$ is calculated as follows.
    Because $V_{r,\delta}$ is a radial function, $\rad[V_{r,0}](\uu,p)$ does not depend on the direction $\uu$.
    Hence, it is sufficient to consider a special case when $(\RR \uu)^\perp = \RR^{m-1}$. Therefore,
    \begin{align}
    \rad[V_{r,0}](\uu,p)
&=\int_{\RR^{m-1}} V_{r,0}( p \uu + y ) \dd \yy, \quad \uu \perp \yy \nonumber \\
&= \int_{\RR^{m-1}} \frac{1}{2}| p \uu + \yy |^2 \ind_{\BB^m(0;r)}(p \uu + \yy) \dd \yy \nonumber \\
    &= \frac{1}{2}  \int_{\BB^{m-1}\left( 0; \sqrt{r^2 - p^2} \right)} \left\{ p^2 + |\yy|^2 \right\} \dd \yy, \label{eq:radV}
    \end{align}
        where the third equation follows by the orthogonality $|p \uu + \yy|^2_m = p^2 + |\yy|^2_{m-1}$ and a geometric consideration as follows:
        \begin{align*}
        \int_{\RR^{m-1}} [\wdot] \ind_{\BB^m(0;r)}(p \uu + \yy) \dd \yy
        &= \int_{\RR^{m-1}} [\wdot] \ind_{\BB^m(-p \uu;r)}(\yy) \dd \yy \\
        &= \int_{\RR^{m-1} \cap {\BB^m(-p \uu;r)}} [\wdot] \dd \yy \\
        &= \int_{{\BB^{m-1}(0; \sqrt{r^2 - p^2})}} [\wdot] \dd \yy.
    \end{align*}
    The first integral in \refeq{radV} is calculated as follows:
    \begin{align}
    \int_{\BB^{m-1}\left( 0; \sqrt{r^2 - p^2} \right)}  p^2 \dd \yy
    &= p^2 \, \vol \left[ \BB^{m-1}(0;\sqrt{r^2 - p^2}) \right] \nonumber \\
    &= \frac{\pi^\frac{m-1}{2}}{2 \Gamma \left( \frac{m-1}{2} +1\right)} p^2 (r^2 - p^2)^\frac{m-1}{2}.
    \end{align}
    The second integral in \refeq{radV} is calculated as follows:
    \begin{align}
    \int_{\BB^{m-1}\left( 0; \sqrt{r^2 - p^2} \right)} |\yy|^2 \dd \yy
    & = \int_{\Sph^{m-2}} \int_{0}^{\sqrt{r^2-p^2}} |\rho \omega|^2 \rho^{m-2} \dd \rho \dd \omega \nonumber \\
    &= \int_{\Sph^{m-2}} \dd \omega \int_{0}^{\sqrt{r^2-p^2}} \rho^{m} \dd \rho  \nonumber \\
    &= \frac{\pi^\frac{m-1}{2}}{(m+1) \Gamma \left( \frac{m-1}{2}\right)} (r^2 - p^2)^\frac{m+1}{2}.
    \end{align}
Hence, by combining the first and second integrals, we have
\begin{align}
    \rad[V_{r,0}](\uu,p) =
    \begin{cases}
    \frac{A_{m-1}}{2(m+1)}(r^2-p^2)^\frac{m-1}{2} \left\{ \frac{2}{m-1} p^2 + r^2  \right\} &  |p| < r \\
    0   & |p| \geq r.
    \end{cases} \label{eq:radVV}
\end{align}

    The ridgelet transform $R_{\rho'} [V_{r,0}]$ is given by
    \begin{align}
    R_{\rho'} [V_{r,0}](\uu/\alpha,\beta/\alpha)=
    \int_{|p| < r} k(p) \overline{\rho'_\alpha(p-\beta )} \dd p,
    \end{align}
    where we define
    \begin{align*}
    k(p) := \rad[V_{r,0}](\uu,p).
    \end{align*}
    Recall that $\rad[V_{r,0}](\uu,p)$ does not depend on the direction $\uu$; thus, the definition of $k$ is reasonable.
    According to \refeq{radVV}, $k$ is a compactly supported bump function.
    Consequently, $k$ is summable; thus, the integral
    \begin{align}
K := \int_{\RR} k(p) \dd p \label{eq:def.K}
    \end{align}
    always exists.
    Recall that the convolution results in smoothing, i.e.,
    \begin{align}
    \int_{|p| < r} k(p) \overline{\rho'_\alpha(p-\beta )} \dd p
    \approx K \overline{\rho'_\alpha(-\beta )}.
    \end{align}

    In summary, we have presented the following:
    \begin{align*}
    R_\rho[\id_{r,0}](\aa, b)
    &=
    -\aa R_{\rho'}[V_{r,0}](\aa, b) \approx -K \aa \overline{\rho'(-b)}. \qedhere
    \end{align*}

\bibliographystyle{plainnat}
\bibliography{./library}

\end{document}